\documentclass{article}

\PassOptionsToPackage{numbers, compress}{natbib}
\usepackage[main, final]{neurips_2026}

\usepackage[utf8]{inputenc} % allow utf-8 input
\usepackage[T1]{fontenc}    % use 8-bit T1 fonts
\usepackage{hyperref}       % hyperlinks
\usepackage{url}            % simple URL typesetting
\usepackage{booktabs}       % professional-quality tables
\usepackage{amsfonts}       % blackboard math symbols
\usepackage{nicefrac}       % compact symbols for 1/2, etc.
\usepackage{microtype}      % microtypography
\usepackage{xcolor}         % colors
\usepackage{hyperref}
\usepackage{url}
\usepackage{amsmath}
\usepackage{amsthm}
\usepackage{amssymb}
\usepackage{wrapfig}
\usepackage{graphicx}
\usepackage{comment}
\usepackage{booktabs}
\usepackage{algorithm}
\usepackage{algpseudocode}

\usepackage{multirow}
\usepackage{float}
\usepackage[table]{xcolor}
\usepackage{colortbl}
\usepackage{markdown}
\usepackage{makecell}
\usepackage{array}
\usepackage{subcaption}
\usepackage[export]{adjustbox}
\usepackage[dvipsnames]{xcolor}
\theoremstyle{definition}

\newcommand{\attackName}{\texttt{LIBRA}}
\newcommand{\attackNameNoColor}{\texttt{LIBRA}}

\newcommand{\protocoltwo}{Protocol \MakeUppercase{\romannumeral 2}}
\newcommand{\best}[1]{{\color{red}{\textbf{#1}}}}
\newcommand{\secondbest}[1]{{\color{blue}{#1}}}

\title{Bayesian Low-Rank Posteriors for \\ Scalable Membership Inference}

\author{%
  \parbox{\textwidth}{\centering
    {Khanh Nguyen\textsuperscript{$\dagger$}}\quad
    {Antonio Emanuele Cinà\textsuperscript{$\S$}}\thanks{The corresponding author}\quad 
    {Andrey Barsky\textsuperscript{$\dagger$}}\quad\\[0.2cm]
    {Alejandra Reinares\textsuperscript{$\dagger$}}\quad
    {Fabio Roli\textsuperscript{$\P\ddagger$}}\quad
    {Dimosthenis Karatzas\textsuperscript{$\dagger$}}\\[2.5pt]
    {\rm
      \textsuperscript{$\dagger$}Computer Vision Center, Universitat Aut\`onoma de Barcelona\\
      \textsuperscript{$\S$} University of Trieste, MIGe, Italy \quad
      \textsuperscript{$\ddagger$} University of Cagliari, DIEE, Italy\\
      \textsuperscript{$\P$} University of Genoa, DIBRIS, Italy\\[2pt]
      \texttt{\{knguyen,abarsky,dimos\}@cvc.uab.es  \quad antonio.cina@units.it \quad alejandra.reinares@autonoma.cat} \quad 
      \texttt{fabio.roli@unige.it}
    }
  }%
}

\usepackage{comment}
\begin{document}

\maketitle

\begin{abstract}
Membership inference attacks (MIAs) aim to determine whether a sample was used during the training of a target model. Strong MIAs rely on reference (or shadow) models to estimate target-conditional score distributions, but training such ensembles is computationally expensive for modern large-scale Vision--Language Models (VLMs). In this work, we propose a scalable alternative that replaces explicit reference-model training with a Bayesian approximation over low-rank adaptation parameters. 
Specifically, we model uncertainty only within the LoRA subspace and construct a posterior from a single training run, from which we sample a diverse set of \emph{virtual reference models}.
We instantiate this approach using Stochastic Weight Averaging Gaussian (SWAG), enabling efficient approximation of target-conditional score distributions at a fraction of the cost of conventional shadow-model ensembles.
We evaluate the resulting attack on three downstream-adapted VLMs across two privacy-sensitive visual question answering tasks, namely Document VQA and Medical VQA. To isolate the effect of pretraining knowledge, we additionally introduce a controlled synthetic dataset.
Our approach outperforms standard reference-based attacks while requiring only a single reference-model fine-tuning run, making membership inference more practical for large VLMs.
\end{abstract}

\section{Introduction}\label{sec:intro}
Vision--Language Models (VLMs)~\cite{liu2023visual,zhu2023minigpt,wang2024qwen2,wu2024deepseek} are commonly adapted from pretrained foundation models through parameter-efficient fine-tuning methods such as LoRA~\cite{hu2022lora,gao2024clip}.
This makes downstream adaptation practical when limited computational resources~\cite{laurenccon2024matters,zanella2024low}, enabling practitioners to specialize VLMs for their tasks.
Such applications often rely on proprietary datasets that are smaller and more domain-specific than the pretraining corpus, containing sensitive information like patient records~\cite{nath2025vila} or confidential business details~\cite{bhatia2024fintral}.
However, adapting to narrower distributions with limited data increases \emph{memorization} risk, encoding sensitive information into the updated parameters.

Membership inference attacks (MIAs)~\cite{shokri2017membership} assess this risk by determining whether a specific data point was used during training, exploiting differences in model behavior between members and non-members such as variations in loss, confidence, or uncertainty~\cite{yeom2018privacy,sablayrolles2019white,watson2022on}.
An adversary can use these signals to infer whether a patient's record~\cite{liu2021slake} or confidential business document~\cite{tobaben2024neurips} was included in downstream fine-tuning.
Despite these risks, research on MIAs for VLMs remains limited. Among existing approaches, strong MIAs rely on multiple \emph{reference models} (or shadow models) trained on datasets that include or exclude the target point~\cite{shokri2017membership,carlini2022membership}, allowing the attacker to conduct target-specific calibrated hypothesis tests. However, training multiple models with the target architecture is computationally prohibitive for large VLMs, limiting their practical applicability.

\begin{figure}[t]
   \centering
   \resizebox{0.99\linewidth}{!}{
   \includegraphics[width=\linewidth]{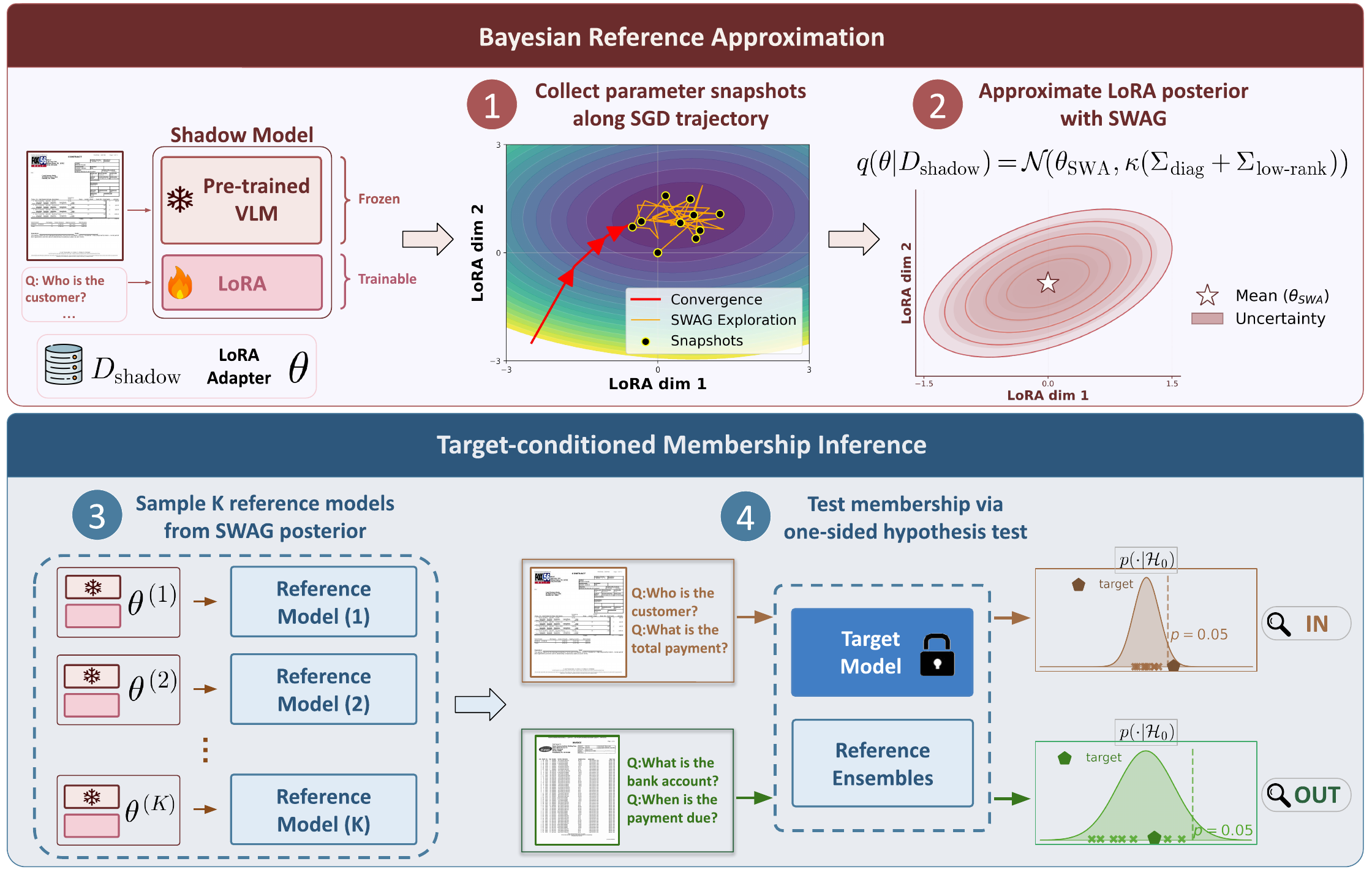}
   }
   \caption{\textbf{Overview of \attackName{}.} \textbf{(1)} LoRA fine-tuning on shadow data; snapshots collected along the SGD trajectory. \textbf{(2)} SWAG fits a Gaussian posterior over the LoRA subspace from these snapshots (contours show a 2D projection). \textbf{(3)} $K$ adapters sampled from the posterior yield \emph{virtual} reference models sharing frozen backbone. \textbf{(4)} Reference scores form an empirical null $p(S \mid \mathcal{H}_0)$; a one-sided test rejects non-membership when the target deviates from the null (direction depends on $S$).}
   \label{fig:teaser}
\end{figure}

In this work, we propose \attackName{} (\emph{Low-rank Membership Inference via Bayesian Reference Approximation}), a scalable alternative that replaces ensembles of independently trained models with a \emph{Bayesian posterior} over LoRA parameters (see Figure~\ref{fig:teaser}). Rather than training multiple models, we perform a single fine-tuning run and construct a posterior that captures parameter variability across training data splits. Sampling from this posterior yields \emph{virtual reference models} that approximate independent training outcomes, enabling efficient target-conditioned calibration without explicit reference-model training. We instantiate this idea using Stochastic Weight Averaging Gaussian (SWAG)~\cite{maddox2019simple}), which generates diverse model samples at negligible cost from a single training trajectory.

We evaluate our approach on three VLMs across two privacy-sensitive Visual Question Answering tasks: Document VQA~\cite{mathew2021docvqa} on structured documents with personal or business information, and Medical VQA~\cite{liu2021slake} with medical images.
To disentangle the role of pretraining knowledge from fine-tuning memorization, we complement real-world datasets with a controlled synthetic dataset.
Our results show that \attackName{} outperforms standard reference-based attacks while requiring only a single trained model, demonstrating that accurate membership inference is feasible for large VLMs.

Our contributions are summarized as follows:
\begin{itemize}
    \item We introduce \attackName{}\footnote{Code is available at \url{https://github.com/khanhnguyen21006/LIBRA}.}, a scalable, reference-free approximation to target-conditional MIAs via Bayesian modeling in the LoRA subspace.
    \item We provide a principled connection between epistemic uncertainty and membership inference in parameter-efficient fine-tuning.
    \item We demonstrate the effectiveness of the proposed approach on realistic multi-modal tasks, achieving improved performance with significantly reduced training cost.
\end{itemize}

\section{Related Work}\label{sec:mia}
Privacy attacks on ML models aim to extract sensitive information about training data from model parameters or outputs~\citep{yeom2018privacy}. 
Among these, \emph{Membership Inference Attacks}~\citep{shokri2017membership,
carlini2022membership,ye2022enhanced} determine whether a target sample $\mathbf{x}$ was used during training and are typically formulated as score-based hypothesis tests. 
Given a target sample $\mathbf{x}$, the attacker computes a membership score $S(\mathbf{x})$ from the model outputs and compares it against a decision threshold $\tau$. These attacks exploit the observation that models often behave differently on training and non-training samples due to memorization or overfitting, for example by assigning higher confidence or lower loss to training data~\cite{yeom2018privacy}.
Under this view, MIAs consist of two components: (i) a membership score $S(\cdot)$ that extracts signal from the model, and (ii) a calibration procedure that maps this score to a decision threshold $\tau$.

\emph{Population-based} methods apply a single global threshold derived from a held-out set of non-members~\citep{yeom2018privacy,salem2018ml,ye2022enhanced,bertran2023scalable}. While computationally efficient, they ignore per-example variability in score distributions and may therefore confound sample difficulty with membership signal~\citep{carlini2022membership,watson2022on}. 

\emph{Reference-model} methods instead estimate target-conditional score distributions by training multiple reference models on auxiliary data and querying them on the target sample~\citep{carlini2022membership,zarifzadeh2023low}. 
We focus on \emph{offline} calibration, which assumes the adversary can train reference models $\{\mathcal{F}_{\theta'_i}\}$ on datasets drawn from $\mathcal{D}$ that exclude the target $\mathbf{x}$. 
LiRA~\citep{carlini2022membership} fits a Gaussian to reference-model scores and applies a one-sided test; RMIA~\citep{zarifzadeh2023low} replaces this with a pairwise likelihood-ratio test against population samples, retaining power with as few as one reference model. 
These methods capture epistemic uncertainty and yield substantially sharper per-example calibration, but require training many reference models and are therefore prohibitive at the scale of modern VLMs. 
Consequently, attacks against VLMs remain almost entirely 
population-based~\citep{zhang2024membership}, and existing approaches extract membership signals directly from the target model~\cite{li2024membership,hu2025membership,yin2026blackbox}. 
However, while effective in practice, their scores lack a properly calibrated null distribution and cannot provide statistical evidence of attack strength.

In this work, we make reference-model calibration tractable for generative VLMs by avoiding curvature computation. 
Closest to our work, \citet{liu2026how} amortize shadow-model costs via a post-hoc Laplace posterior over model parameters; like Laplace-based Bayesian LoRA more broadly~\citep{yang2024bayesian}, the approach is restricted to classification tasks with a fixed output space, where the requisite curvature estimates remain tractable. 
We instead adopt SWAG~\citep{maddox2019simple}, which constructs a Gaussian posterior from the SGD trajectory without requiring curvature computation. 
SWAG is agnostic to the output structure, making reference-based MIA tractable for generative VLMs at negligible additional cost.

\section{Methodology}
We present \attackName{} (Figure~\ref{fig:teaser}) through its threat model (Section~\ref{sec:threat_model}), target-conditional hypothesis testing formulation (Section~\ref{sec:hypothesis_testing}), and SWAG approximation of LoRA reference ensembles (Section~\ref{sec:bayesian_lora}).

\subsection{Threat Model}
\label{sec:threat_model}

\textbf{Downstream adaptation setting.}
We focus on the downstream adaptation of VLMs, where models are fine-tuned on small domain-specific datasets containing sensitive information. 
Our threat model, therefore, targets privacy leakage introduced during downstream adaptation rather than memorization inherited from pretraining.
At the same time, since full fine-tuning of VLMs is computationally prohibitive, practitioners commonly adopt parameter-efficient methods such as Low-Rank Adaptation (LoRA)~\cite{hu2022lora}. 
In \autoref{app:preliminaries}, we provide a brief overview of the LoRA mechanism.

\textbf{Adversary Knowledge.}
We consider a realistic \emph{grey-box} adversary with access to the pretrained VLM, which is publicly available as a foundation model and reused for downstream applications via PEFT adaptation. This assumption reflects current practice, where pretrained checkpoints are openly distributed through repositories such as HuggingFace.
The adversary has no access to the private fine-tuning dataset $\mathrm{D}_{\mathrm{train}}$, but is assumed to know the low-rank adaptation training algorithm. We relax this assumption in Section~\ref{sec:cross_transfer}. 
In addition, the adversary has access to a limited shadow dataset $\mathrm{D}_{\mathrm{shadow}} \sim \mathcal{D}$, drawn from the same underlying distribution as $\mathrm{D}_{\mathrm{train}}$ but disjoint from it.
We further assume realistic computational constraints under which repeatedly retraining large VLMs is infeasible, ruling out attacks that require training hundreds of shadow models~\cite{carlini2022membership}.
The adversary therefore seeks efficient attacks that extract meaningful membership signals within a limited computational budget.

\subsection{Membership Inference as Hypothesis Testing in \attackNameNoColor}
\label{sec:hypothesis_testing}
Membership inference attacks~\cite{ye2022enhanced, carlini2022membership,bertran2023scalable, zarifzadeh2023low} are commonly formulated as a hypothesis test that determines whether a target sample $\mathbf{x}$ belongs to the training dataset $\mathrm{D}_{\mathrm{train}}$:
\begin{equation}
\label{eq:hypothesis_test}
    \mathcal{H}_0: \; \mathbf{x} \sim \mathcal{D} \setminus \mathrm{D}_{\mathrm{train}} \;\;(\text{non-member}), \qquad
    \mathcal{H}_1: \; \mathbf{x} \sim \mathrm{D}_{\mathrm{train}} \;\;(\text{member}).
\end{equation}
where $\mathbf{x}$ denotes denotes the target of the membership inference. The test must be conditioned on the specific target $\mathbf{x}$, since score distributions are highly heterogeneous across targets~\cite{carlini2022membership}—a heterogeneity amplified in VLMs by joint variation in visual complexity and question diversity. We therefore characterize the \emph{target-conditional} null $p(S \mid \mathcal{H}_0(\mathbf{x}))$ and reject $\mathcal{H}_0(\mathbf{x})$ when $S(\mathbf{x})$ exceeds a target-dependent threshold $\tau_\alpha(\mathbf{x})$. Under the true null distribution, this threshold is chosen to yield a per-target false positive rate of $\alpha$~\cite{ye2022enhanced,bertran2023scalable}:
\begin{equation}\label{eq:fpr_guarantee}
    \mathbb{E}_{p(S \mid \mathcal{H}_0(\mathbf{x}))}\!\left[\mathbf{1}\!\left[S(\mathbf{x}) \geq \tau_\alpha(\mathbf{x})\right]\right] = \alpha.
\end{equation}
Reliable estimation of $\tau_\alpha(\mathbf{x})$ requires resolving the tails of $p(S \mid \mathcal{H}_0(\mathbf{x}))$, which demands many reference samples per target. 
We adopt a Gaussian approximation~\cite{carlini2022membership}, summarizing the null by its first two moments $\mu(\mathbf{x}), \sigma(\mathbf{x})$ and thresholding the standardized score:
\begin{equation}
\label{eq:z_score}
    z(\mathbf{x}) = \frac{S(\mathbf{x}) - \mu(\mathbf{x})}{\sigma(\mathbf{x})}
\end{equation}
 
We compare this score against a global threshold $\tau$, shared across all target samples and varied over $\mathbb{R}$ to trace the TPR--FPR curves in Section~\ref{sec:results}.
Given a target sample $\mathbf{x}$ consisting of an image $\mathbf{v}$, a text input $\mathbf{q}$, and a text output $\mathbf{a}$, we define the membership score via the auto-regressive likelihood:
\begin{equation}
\label{eq:softmax_probability}
    p_\theta(\mathbf{a} \mid \mathbf{v}, \mathbf{q}) = \prod_{t=1}^{T} p_\theta\!\left(a_t \mid \mathbf{z}_v, \mathbf{q}, \mathbf{a}_{<t}\right),
\end{equation}
where $\mathbf{z}_v$ denotes the visual tokens produced by the VLM and each next-token probability uses a softmax with temperature $\beta$.
Since softmax probabilities are poorly approximated by a Gaussian, we apply a logit transform~\cite{carlini2022membership} $S(\mathbf{x}) = \log(p / (1-p))$, which approximates Gaussian output (Figure~\ref{fig:scaled_logit_dist}).

\subsection{Motivations}
\label{sec:motivations}
As reviewed in Section~\ref{sec:mia}, conditional calibration of $p(S \mid \mathcal{H}_0(\mathbf{x}))$ follows two paradigms: population-based and reference-model methods, modeling aleatoric and epistemic uncertainty respectively. We argue that reference-model calibration is better aligned with the counterfactual semantics of Eq.~\ref{eq:hypothesis_test}, and design \attackName\ to preserve this alignment at a cost tractable for VLMs.

\paragraph{Why reference models.}
The null $\mathcal{H}_0(\mathbf{x})$ is inherently counterfactual: it is the distribution of $S(\mathbf{x})$ under models that \emph{never saw} $\mathbf{x}$ during training. Reference-model ensembles realize this counterfactual directly, each model providing an independent draw from the space of models trained without $\mathbf{x}$. Population-based methods, in contrast, evaluate non-member scores under \emph{the target model itself}, whose training included $\mathrm{D}_{\mathrm{train}}$; any statistic derived from the target inherits this exposure, and conditioning on features of $\mathbf{x}$ cannot disentangle the membership signal from example difficulty.

\begin{figure}[htbp]
    \centering
    \begin{subfigure}[t]{0.5\textwidth}
        \vspace{0pt}
        \centering
        \includegraphics[width=\linewidth]{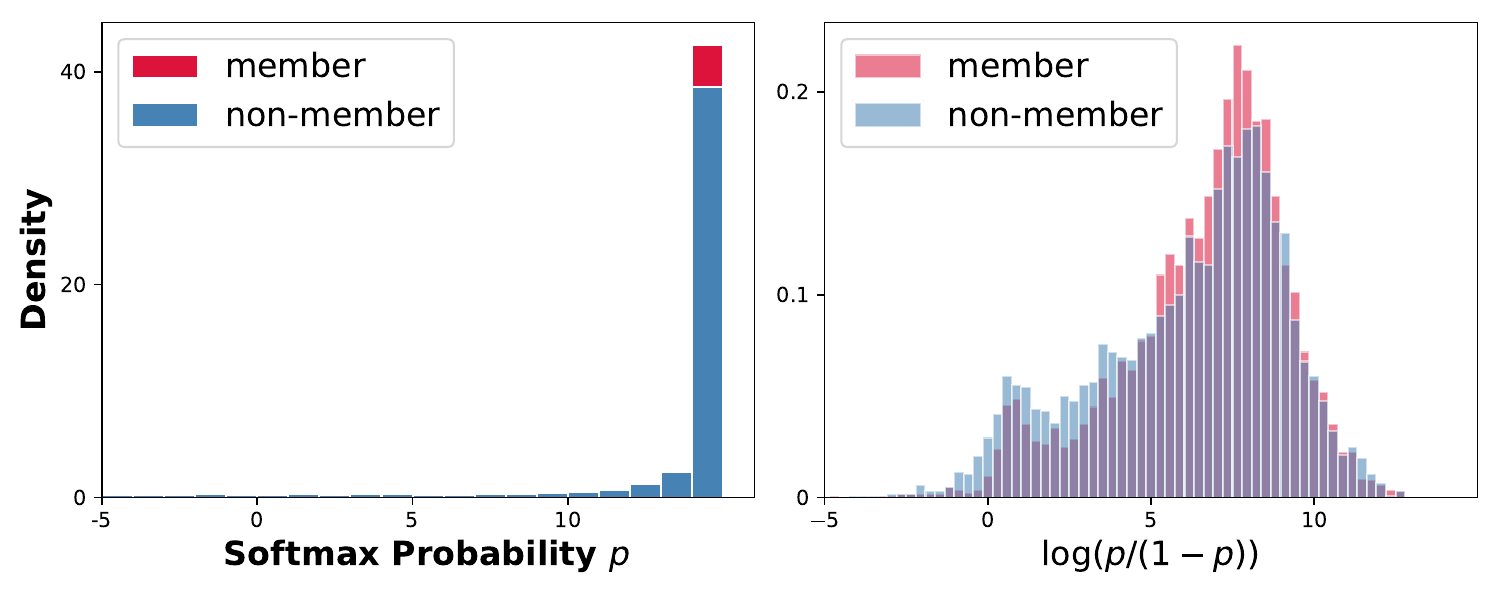}
        \caption{\textbf{Attack score distributions on Qwen2-VL fine-tuned for DocVQA}. Raw softmax probability (\textit{left}) and the scaled-logit transform~\cite{carlini2022membership}, which is approximately Gaussian (\textit{right}), are shown for members and non-members. Both distributions exhibit substantial overlap between the two classes, indicating that membership cannot be reliably inferred from the raw score.}
        \label{fig:scaled_logit_dist}
    \end{subfigure}
    \hfill
    \begin{subfigure}[t]{0.47\textwidth}
        \vspace{0pt}
        \centering
        \includegraphics[width=\linewidth]{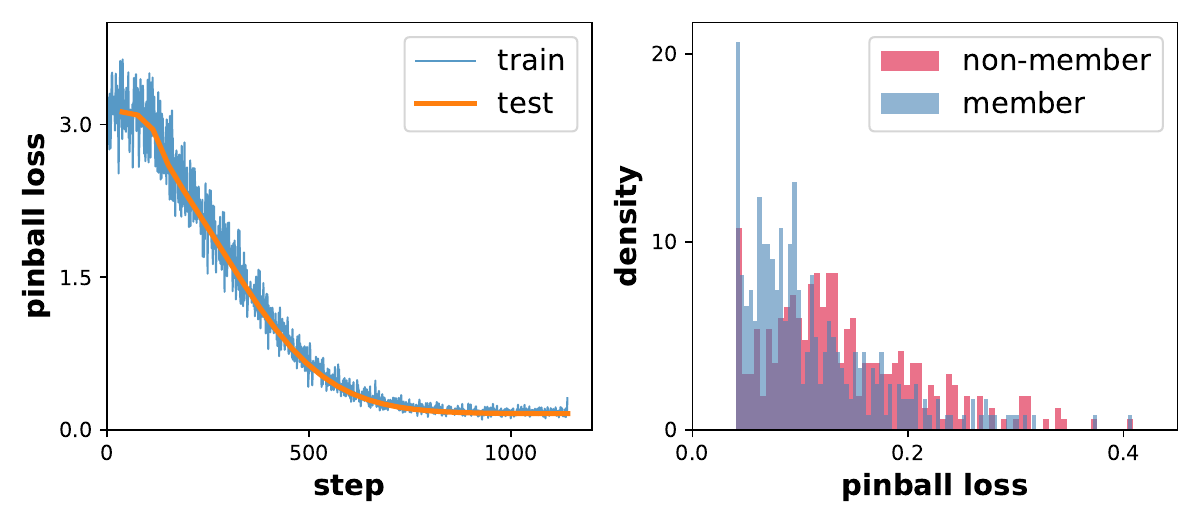}
        \vspace{-1.4em}
        \caption{\textbf{Q-MIA pinball-loss diagnosis on Qwen2-VL fine-tuned for DocVQA.} \textit{(Left)} The quantile regressor trains cleanly without overfitting. \textit{(Right)} Yet pinball-loss distributions of members and non-members overlap heavily under the trained regressor, showing that the regressor has absorbed the membership signal rather than exposing it as a gap.}
        \label{fig:quantile_regressor}
    \end{subfigure}
    \label{fig:qwen2vl_score_distribution}
            \vspace{-0.5em}
\end{figure}

We illustrate this confounding on Qwen2-VL~\cite{wang2024qwen2} fine-tuned for DocVQA. Following~\cite{tito2024privacy}, we synthesize business-invoice images and QA pairs with controlled stylistic variation, ensuring high visual similarity between members and non-members (Figure~\ref{fig:docvqa_synthetic_examples}) so that membership becomes the dominant source of distributional difference. We then train a Q-MIA~\cite{bertran2023scalable} quantile regressor $\hat{q}_\alpha(\mathbf{x})$ on non-member scores by minimizing the pinball loss
$\mathcal{L}_\alpha(s, \hat{q}) = \max\bigl(\alpha(s - \hat{q}),\, (\alpha - 1)(s - \hat{q})\bigr)$
averaged over $\alpha \in (0, 1)$, with $s = S(\mathbf{x})$ and $\hat{q} = \hat{q}_\alpha(\mathbf{x})$. Figure~\ref{fig:quantile_regressor} (left) shows clean training: pinball loss converges and validation tracks training. Yet the right panel exposes the failure: member and non-member loss distributions overlap heavily. A calibrated regressor should under-predict member scores, leaving the membership signal as a residual; here it generalizes to members and absorbs that signal. The features used to model aleatoric variability are entangled with membership, so a well-trained regressor partially explains the signal away. Section~\ref{sec:results} confirms that reference-based attacks dominate population-based baselines at low FPR on LoRA fine-tuned VLMs.

\paragraph{Why Bayesian LoRA.}
Reference-based calibration is statistically principled but computationally expensive, as attacks such as LiRA~\citep{carlini2022membership} rely on multiple reference models, each requiring a separate fine-tuning run over the LoRA adapters.
We resolve this tension by replacing the reference ensemble with a \emph{ Bayesian posterior over the LoRA adapters of a single reference model}, from which reference models can be sampled at negligible marginal cost.
Specifically, we adopt Stochastic Weight Averaging Gaussian (SWAG)~\citep{maddox2019simple}, which approximates the adapter posterior by a Gaussian fitted from SGD iterates along a single fine-tuning trajectory. Since only the low-rank adapters are stochastic while the frozen backbone is shared across samples, drawing $K$ posterior samples costs $K$ low-rank additions per adapted layer, making reference-based MIA tractable for LoRA fine-tuned VLMs.

\subsection{Bayesian Low-Rank Adaptation for MIAs}
\label{sec:bayesian_lora}
We now instantiate the Bayesian reference ensemble motivated in Section~\ref{sec:motivations}. 

\textbf{Bayesian LoRA for reference modeling.}
Formally, let $\boldsymbol{\theta} \in \mathbb{R}^p$ denote the set of trainable LoRA parameters introduced during downstream adaptation, while the pretrained backbone remains frozen and is shared across all reference models.
Rather than point-estimating $\boldsymbol{\theta}$ on a shadow dataset $\mathrm{D}_{\mathrm{shadow}}$ disjoint from $\mathrm{D}_{\mathrm{train}}$, we treat the adapter as a random variable with posterior $q(\boldsymbol{\theta} \mid \mathrm{D}_{\mathrm{shadow}})$. 
Each sample from this posterior corresponds to one plausible fine-tuning outcome. Hence, repeatedly sampling from $q$ yields a collection of \emph{virtual reference models} that approximate independently trained shadow models. In particular, $K$ such samples provide the reference ensemble required by Eq.~\ref{eq:z_score}, while requiring only a single fine-tuning run.
We approximate the posterior $q$ with a Gaussian distribution $\mathcal{N}(\boldsymbol{\mu}, \Sigma)$, which in turn induces an approximately Gaussian conditional null distribution $p(S \mid \mathcal{H}_0(\mathbf{x}))$. However, materializing the full $p \times p$ covariance matrix is infeasible at VLM scale. We therefore parameterize $\Sigma$ implicitly using moments of the SGD trajectory through SWAG~\cite{maddox2019simple, onal2024gaussian}.

\paragraph{SWAG posterior over LoRA subspace.}
Starting from a converged adapter $\boldsymbol{\theta}^\star$ obtained by tuning the VLM on $\mathrm{D}_{\mathrm{shadow}}$, we continue training with a constant learning rate and periodically snapshot the adapter parameters during a collection phase. Let $\boldsymbol{\theta}_{1}, \boldsymbol{\theta}_{2}, \dots$ denote the resulting sequence of SGD iterates. SWAG maintains two running statistics:
\begin{equation}
\bar{\boldsymbol{\theta}}_t = \frac{1}{t}\sum_{j=1}^{t}\boldsymbol{\theta}_j,
\qquad
\overline{\boldsymbol{\theta}^2}_t = \frac{1}{t}\sum_{j=1}^{t}\boldsymbol{\theta}_j \odot \boldsymbol{\theta}_j,
\end{equation}
yielding the diagonal covariance $\Sigma_{\mathrm{diag}} = \mathrm{diag}\bigl(\overline{\boldsymbol{\theta}^2} - \bar{\boldsymbol{\theta}} \odot \bar{\boldsymbol{\theta}}\bigr)$, where $\odot$ denotes the Hadamard product. 
To capture off-diagonal structure without forming $\Sigma$, SWAG additionally maintains a sliding buffer of the $M$ most recent centered deviations $\mathbf{d}_s = \boldsymbol{\theta}_s - \bar{\boldsymbol{\theta}}_s$ as columns of $\hat{D} \in \mathbb{R}^{p \times M}$, giving a low-rank component $\Sigma_{\mathrm{low-rank}} = \tfrac{1}{M-1}\hat{D}\hat{D}^{\top}$. The adapter posterior is:
\begin{equation}
q(\boldsymbol{\theta} \mid \mathrm{D}_{\mathrm{shadow}})
= \mathcal{N}\!\left(\bar{\boldsymbol{\theta}},\; \kappa\bigl(\Sigma_{\mathrm{diag}} + \Sigma_{\mathrm{low-rank}}\bigr)\right).
\label{eq:swag_posterior}
\end{equation}
where ${\kappa} > 0$ is a scale factor that controls the spread of the posterior. Note that $\Sigma$ is never explicitly instantiated, since posterior sampling operates only through the diagonal standard deviations $\Sigma_{\mathrm{diag}}^{1/2}$ and the deviation matrix $\hat{D}$. These require $\mathcal{O}(p)$ and $\mathcal{O}(pM)$ storage, respectively, avoiding the $\mathcal{O}(p^2)$ cost of storing the full covariance matrix.

\paragraph{Sampling reference VLMs.}
At attack time, we draw $K$ samples from the posterior in Eq.~\ref{eq:swag_posterior},
\begin{equation}
\boldsymbol{\theta}^{(k)}
= \bar{\boldsymbol{\theta}}
+ \sqrt{\kappa}\,\Sigma_{\mathrm{diag}}^{1/2}\,\mathbf{z}_1^{(k)}
+ \sqrt{\tfrac{\kappa}{(M-1)}}\,\hat{D}\,\mathbf{z}_2^{(k)},
\qquad \mathbf{z}_1^{(k)} \sim \mathcal{N}(0, I_p),\;\; \mathbf{z}_2^{(k)} \sim \mathcal{N}(0, I_M) 
\label{eq:swag_sample}
\end{equation}
and materialize the $k$-th reference VLM by reattaching $\boldsymbol{\theta}^{(k)}$ to the frozen backbone; only the low-rank adapters differ across the $K$ references. In our experiments (Section~\ref{sec:experimental_setup}), the number of trainable adapter parameters is approximately $p \approx 10^{7}$, compared to a ${\sim}10^{9}$-parameter backbone, so generating the full reference ensemble incurs only a small fraction of the cost of a single VLM forward pass.\\
For each target $\mathbf{x}$, the $K$ sampled VLMs produce scores $\{S_k(\mathbf{x})\}_{k=1}^{K}$ forming a Monte Carlo approximation to the conditional null $p(S \mid \mathcal{H}_0(\mathbf{x}))$; increasing $K$ refines this estimate without altering the posterior. Section~\ref{sec:hypothesis_testing} details how this empirical null is calibrated and combined with the target score into the final test statistic. Algorithm~\ref{alg:algorithm} presents a pseudo-code for our attack \attackName.

\paragraph{Fidelity of the counterfactual.}
Our SWAG ensemble serves the same purpose as the shadow models used in offline reference-based MIAs~\cite{carlini2022membership, zarifzadeh2023low}, providing reference models trained on $\mathrm{D}_{\mathrm{shadow}}$ and reused across targets.
The main difference is that, instead of training each reference model independently, we generate all $K$ references by sampling from a single Gaussian posterior centered at $\bar{\boldsymbol{\theta}}$.
The low-dimensional space ensures that independently trained adapters remain within a comparable parameter region, enabling our posterior to capture the score-level variability needed for calibration.
Despite this approximation, Section~\ref{sec:results} demonstrates that the resulting ensemble achieves state-of-the-art membership inference performance on LoRA fine-tuned VLMs with a small fraction of the computational cost of conventional shadow-model ensembles.
Section~\ref{sec:fidelity} validates this fidelity: the posterior means agree strongly with reference-ensemble means, while the scale factor $\kappa$ controls agreement on the variances, confirming that SWAG samples capture the required score-level statistics.

\section{Experimental Setup}
\label{sec:experimental_setup}
We focus our experiments on Visual Question Answering settings where the image is the primary source of sensitive information, as in document or Medical VQA.
Personally identifiable information (PII) embedded in images is substantially harder to audit and redact than textual PII, making image-level leakage the principal concern. 
We define two attack scenarios based on the unit of membership:
\textbf{\emph{(1) Image-level MIA}}, where the adversary determines whether a given image was in the fine-tuning set. We consider two sub-settings~\cite{nguyen2025docmia,hu2025membership}: \emph{multi-sample}, where the adversary holds all ground-truth QA pairs for the target image, and \emph{image-only}, where only the image is available and QA pairs must be synthesized. We evaluate both on \textbf{PFL-DocVQA}~\citep{tobaben2024neurips}, a privacy benchmark of business invoices.
\textbf{\emph{(2) Sample-level MIA}}, where membership is defined per $(v, q, a)$ tuple, so the same image with different questions constitutes distinct samples~\cite{shokri2017membership,carlini2022membership}. We evaluate this on \textbf{SLAKE}~\citep{liu2021slake}, a Medical VQA benchmark.
Full implementation details for our experiments are deferred to Appendix~\ref{app:impl_details}.

\textbf{Data synthesis.}
For DocVQA experiments, we construct a subset of evaluation samples via an independent synthesis pipeline that produces documents and question-answer pairs in the visual and linguistic style of PFL-DocVQA; representative examples are in Figure~\ref{fig:docvqa_synthetic_documents}. Synthetic data serves two purposes: (i)~it guarantees that evaluated samples were never seen during pre-training, eliminating data contamination while keeping distribution shift minimal; and (ii)~it mirrors realistic deployment, where fine-tuning data is proprietary rather than public. Full details are deferred to Appendix~\ref{app:dataset}.

\textbf{Target VLMs.}
We evaluate three open-weight VLMs with distinct architectures: \textbf{Qwen2-VL}~\cite{wang2024qwen2}, which encodes images at native resolution; \textbf{Idefics2}~\cite{laurenccon2024matters}, which compresses visual tokens through a perceiver resampler; 
and \textbf{DeepSeek-VL2}~\cite{wu2024deepseek}, which pairs a vision encoder with a MoE language backbone.
This variety lets us test whether privacy leakage is a model-specific artifact or a systematic property of fine-tuned VLMs. 
More details on model finetuning are provided in Appendix~\ref{app:target_model}.

\textbf{Baselines}
We focus on the offline attack setting and compare it against representative MIAs covering the two dominant paradigms. 
As for population-based methods, we include \emph{Attack-P}~\cite{ye2022enhanced}, which uses a global threshold $\tau$, and \emph{QMIA}~\cite{bertran2023scalable}, which predicts per-sample thresholds through quantile regression.
Reference-based methods include \emph{LiRA}~\cite{carlini2022membership}, which fits a Gaussian to reference scores, and \emph{RMIA}~\cite{zarifzadeh2023low}, which considers a pair-wise likelihood ratio between the target sample and a population set, again in an offline setting.
We also include two score-function attacks originally developed for LLMs: \emph{Min-K\% Prob}~\cite{shi2024detecting} and \emph{MaxRényi-K\%}~\cite{li2024membership}. Full descriptions are in Appendix~\ref{app:baselines}.

\textbf{Evaluation Protocol}
\label{sec:evaluation_protocol}
We evaluate MIAs on $\mathrm{D}_{\mathrm{train}} \cup \mathrm{D}_{\mathrm{test}}$, balanced between members and non-members. For each target $\mathbf{x}$, we compute a score $s(\mathbf{x})$ and sweep a single global threshold to trace the ROC curve. This matches how LiRA and RMIA are reported in practice: although their scores are per-sample likelihood ratios, running them as proper hypothesis tests at level $\alpha$ would require estimating the per-sample null tail, which demands far more reference models than is feasible. Population-based attacks avoid this cost, since their null is fit to a large held-out population; we report per-sample test results in Appendix~\ref{app:results}. Evaluation metrics are AUROC, and TPR at $3\%$ and $5\%$ FPR.

\section{Results}
\label{sec:results}
We report multi-sample and image-only MIA results on PFL-DocVQA across three target VLMs. Sample-level MIA results follow the same trends and are deferred to Appendix~\ref{app:results}.

\textbf{Multi-sample MIA.}
\begin{figure}[tb]
    \centering
    \hfill
    \begin{subfigure}{0.325\textwidth}
        \centering
        \includegraphics[width=\linewidth]{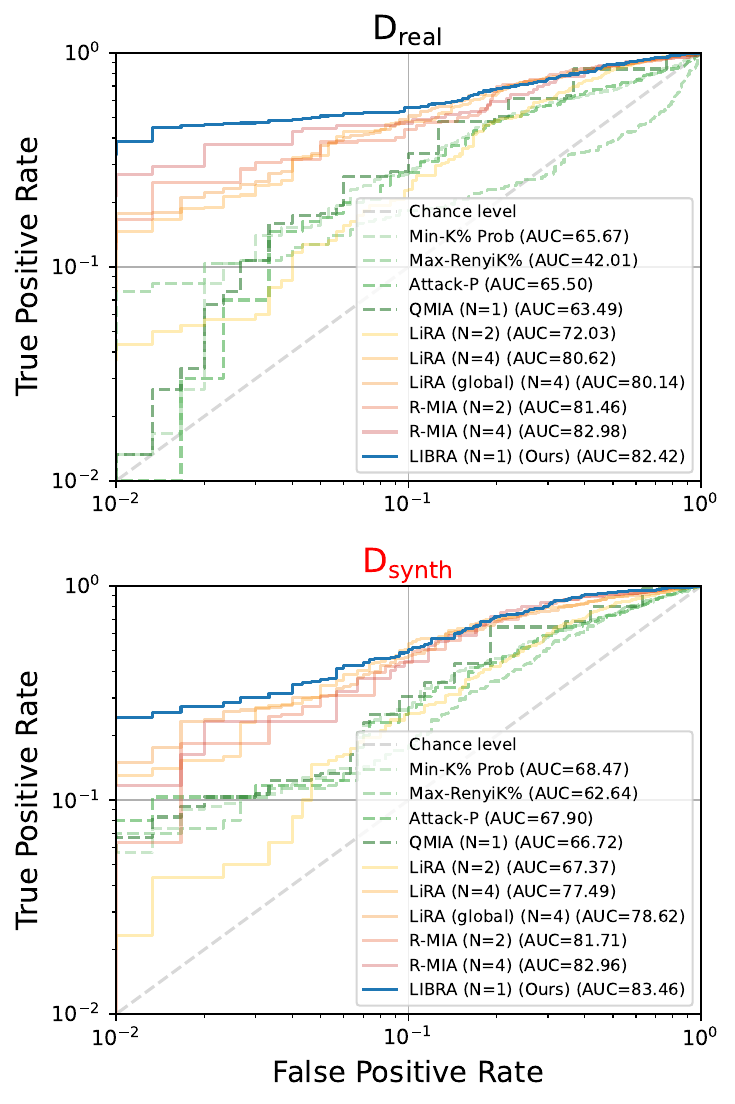}
        \caption*{Idefics2}
    \end{subfigure}
    \begin{subfigure}{0.325\textwidth}
        \centering
        \includegraphics[width=\linewidth]{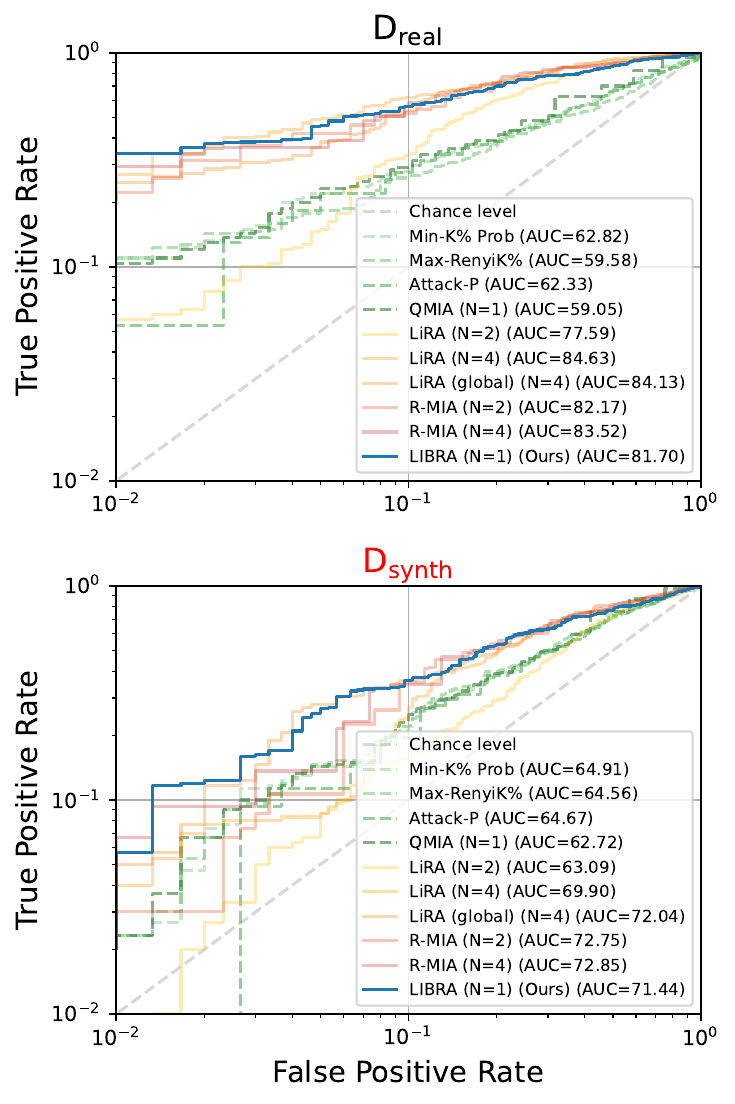}
        \caption*{Qwen2-VL}
    \end{subfigure}
    \hfill
    \begin{subfigure}{0.325\textwidth}
        \centering
        \includegraphics[width=\linewidth]{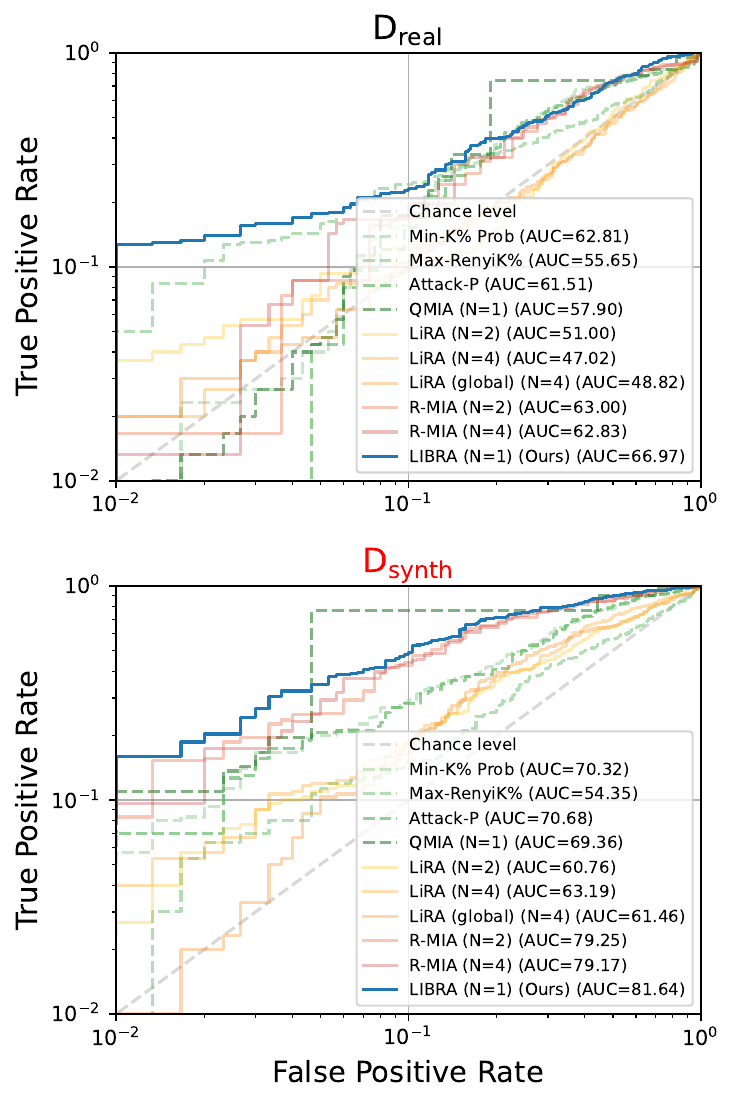}
        \caption*{DeepSeek-VL2}
    \end{subfigure}
    \caption{\textbf{Multi-sample MIA on DocVQA.} ROC curves (log scale) comparing \attackName{} to existing MIAs. Evaluation is split across the real ($\mathrm{D}_{\mathrm{real}}$) and synthetic ($\mathrm{D}_{\mathrm{\textcolor{red}{synth}}}$) subsets of $\mathrm{D}_{\mathrm{test}}$. $N$ denotes the number of trained reference models. Results for \attackName{} use an ensemble of $K=15$ reference models sampled from the posterior. TPR@ 3\% and 5\% FPR readout are provided in Table~\ref{tab:result_multi_sample_docvqa}.
    %LiRA* represents LiRA with fixed variance estimate.
    }
    \label{fig:result_multi_sample_docvqa}
\end{figure}
Since each target document is associated with multiple questions, we average membership scores across all QA pairs per document. Figure~\ref{fig:result_multi_sample_docvqa} and Table~\ref{tab:result_multi_sample_docvqa} report ROC curves and TPR at $3\%$ and $5\%$ FPR. Reference-model attacks consistently dominate population-data baselines across all models and subsets, consistent with the motivation in Section~\ref{sec:motivations}. Despite training only one reference model, \attackName{} achieves the best or second-best TPR in all configurations. Gains are particularly pronounced on DeepSeek-VL2, where \attackName{} reaches $14.0\%$ TPR at $3\%$ FPR on $\mathrm{D}_{\mathrm{real}}$ versus $5.3\%$ for the strongest baseline. On Idefics2 and Qwen2-VL, it remains competitive with LiRA and RMIA using up to four models, with consistent improvements at low FPR.

\textbf{Image-only MIA.}
In this setting, the attacker has access only to the target document image and must synthesize its own QA pairs. This reflects a more realistic threat model but requires the attack to remain robust under mild distribution shifts, as the generated questions are rephrasings rather than exact matches to those seen during training.
\begin{figure}[t]
    \centering

    \begin{subfigure}{0.325\textwidth}
        \centering
        \includegraphics[width=\linewidth]{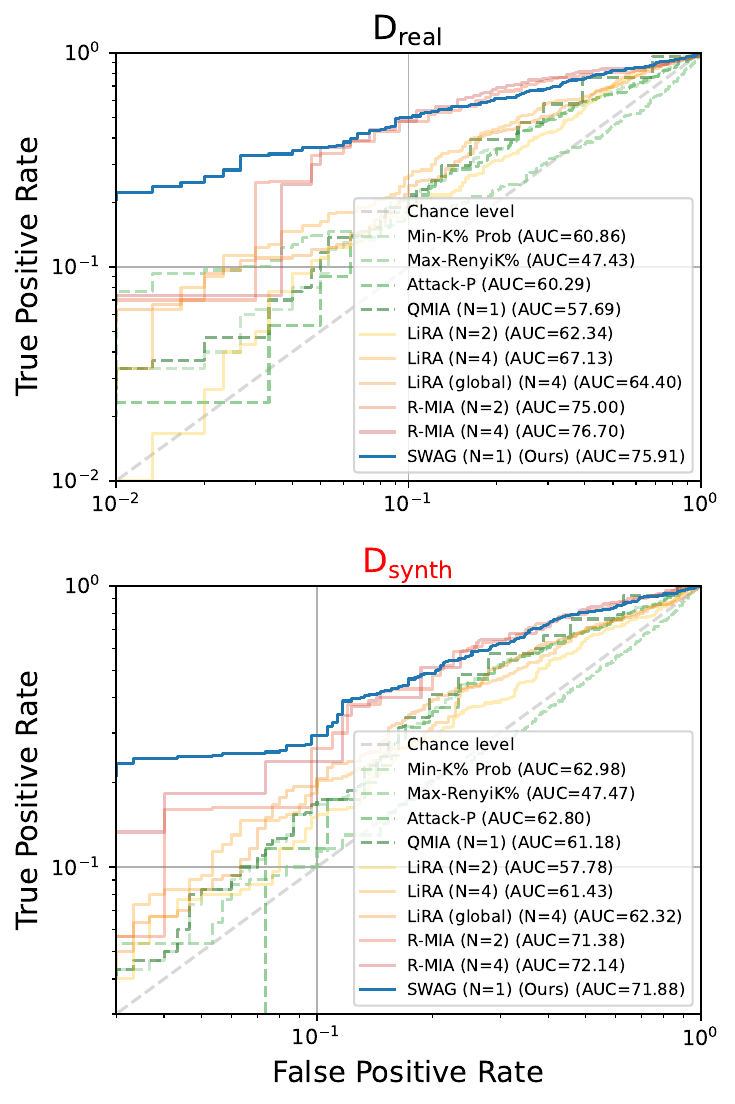}
        \caption*{Idefics2}
    \end{subfigure}
    \hfill
    \begin{subfigure}{0.325\textwidth}
        \centering
        \includegraphics[width=\linewidth]{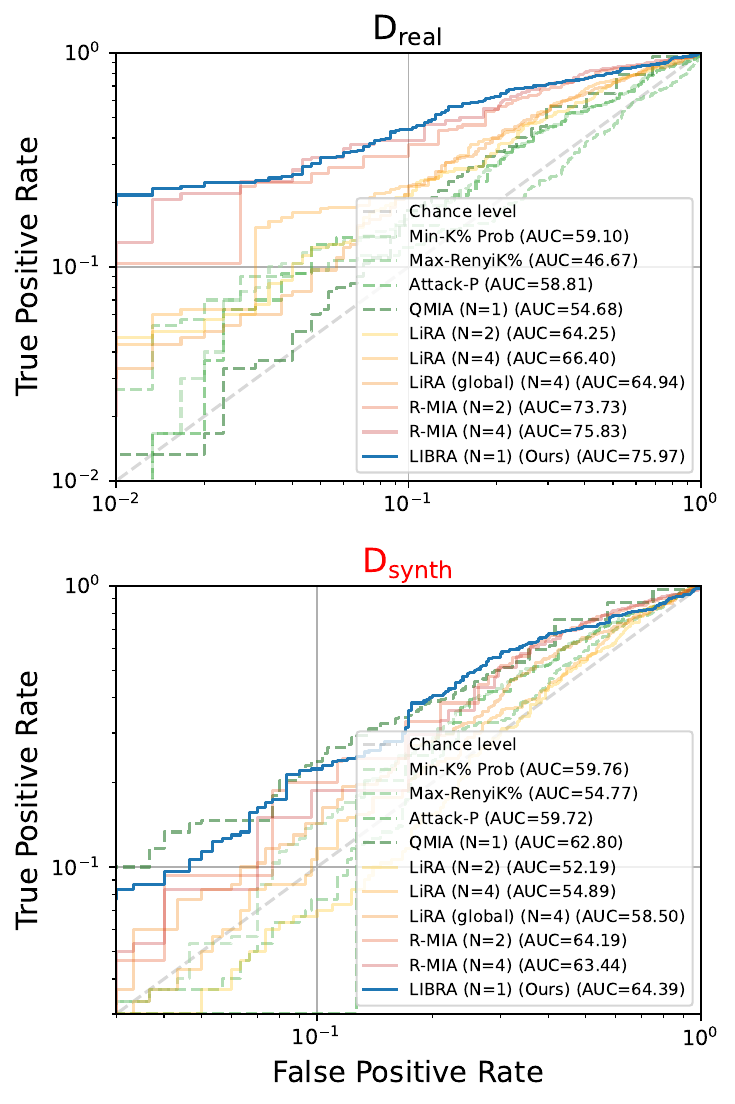}
        \caption*{Qwen2-VL}
    \end{subfigure}
    \hfill
    \begin{subfigure}{0.325\textwidth}
        \centering
        \includegraphics[width=\linewidth]{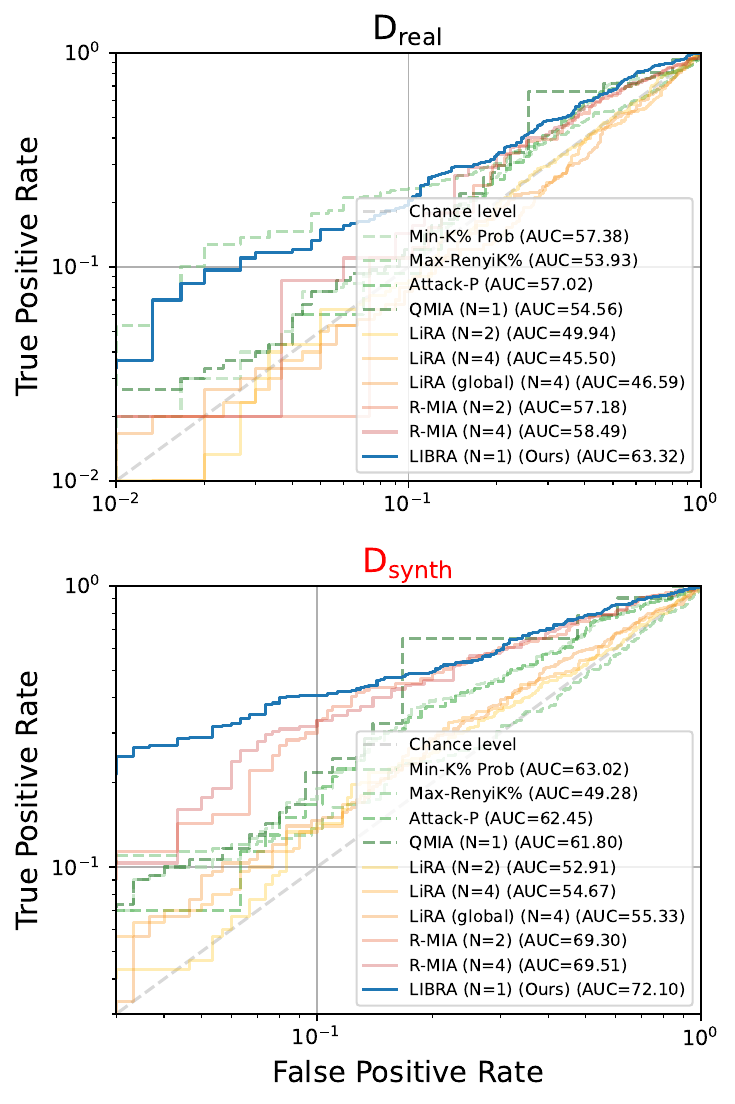}
        \caption*{DeepSeek-VL2}
    \end{subfigure}
    \caption{\textbf{Image-only MIA on DocVQA}. ROC curves (log scale) comparing \attackName{}  to existing MIAs on both $\mathrm{D}_{\mathrm{real}}$ and $\mathrm{D}_{\mathrm{\textcolor{red}{synth}}}$ subsets. Results for \attackName{} use an ensemble of $K=15$ reference models sampled from the posterior. TPR@ 3\% and 5\% FPR readout are provided in Table~\ref{tab:result_image_only_docvqa}.}
    \label{fig:result_image_only_docvqa}
\end{figure}
Table~\ref{tab:result_image_only_docvqa} reports TPR at $3\%$ and $5\%$ FPR, and Figure~\ref{fig:result_image_only_docvqa} shows the ROC curves.
As expected, all attacks degrade relative to the multi-sample setting. 
Even so, \attackName{} achieves the best or second-best TPR in every configuration with a single reference model, while reference-based baselines such as LiRA collapse to near-random performance. 
The gap is most striking on Idefics2, where \attackName{} significantly outperforms the strongest baseline regarding TPR at $3\%$ FPR on $\mathrm{D}_{\mathrm{real}}$ ($33.0\%$ vs.\ $11.3\%$) and on $\mathrm{D}_{\mathrm{synth}}$ ($21.0\%$ vs.\ $13.3\%$). This robustness to query-side distribution shift broadens the practical attack surface, enabling privacy auditing even when the adversary holds only the document image.

\section{Analysis}
\label{sec:analysis}
We analyze the validity of the core approximation from \attackName{} and investigate the factors influencing its performance. Unless otherwise noted, all experiments use the multi-sample MIA setting on DocVQA with Qwen2-VL as the target model.

\subsection{Validity of the SWAG approximation}
\label{sec:fidelity}
\begin{figure}[htbp]
    \centering
    \begin{subfigure}[t]{0.49\textwidth}
        \vspace{0pt}
        \centering
        \includegraphics[width=\linewidth]{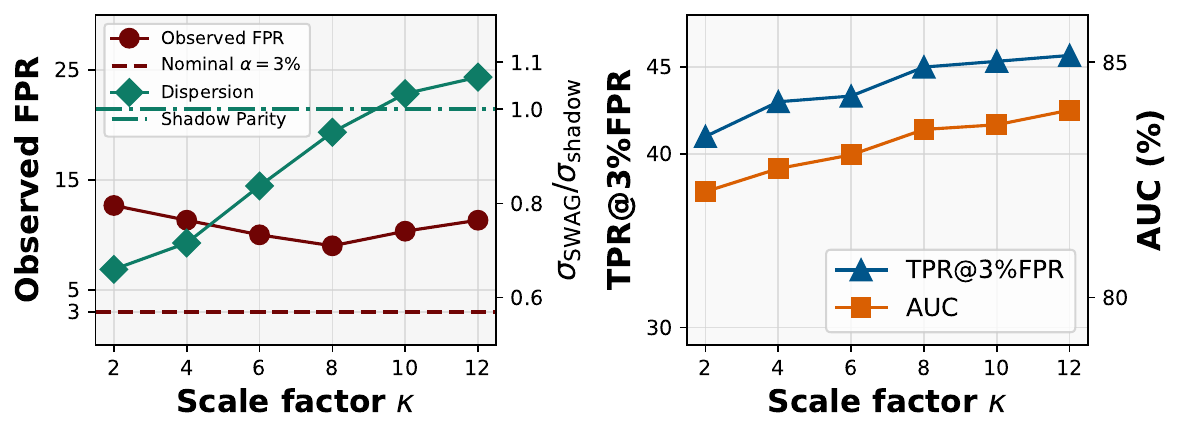}
        \caption{\textbf{Effect of SWAG posterior scale factor $\kappa$} (Qwen2-VL, per-sample threshold at nominal $\alpha{=}3\%$; validation set). \textbf{\textit{(left)} Calibration.} Observed FPR (left axis) and dispersion ratio $\sigma_{\text{SWAG}}/\sigma_{\text{shadow}}$ (right axis). Dashed lines mark nominal $\alpha$ and shadow-ensemble parity. \textbf{\textit{(right)} Attack power.} AUC (left axis) and TPR@3\%FPR (right axis). Scaling the posterior improves calibration without sacrificing power.}
        \label{fig:ablate_kappa}
    \end{subfigure}
    \hfill
    \begin{subfigure}[t]{0.49\textwidth}
        \vspace{0pt}
        \centering
        \includegraphics[width=\linewidth]{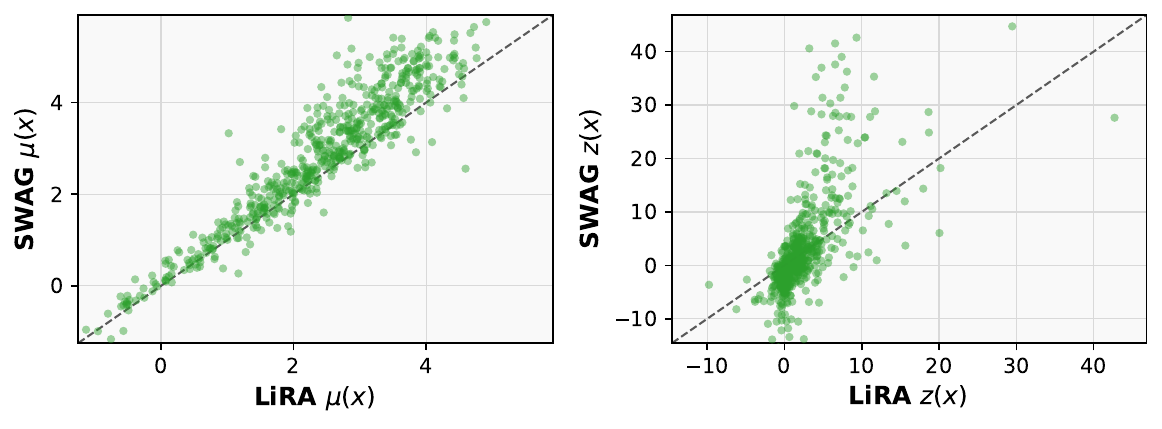}
        \caption{\textbf{Per-target agreement between SWAG and LiRA null estimates.} Each point is one target; dashed line marks perfect agreement. \textbf{Left:} null means $\mu(\mathbf{x})$ agree closely (Spearman $\rho = 0.926$), confirming that both methods recover the same per-target difficulty structure. \textbf{Right:} standardized scores $z(\mathbf{x})$ correlate well ($\rho = 0.771$); the upward tilt reflects SWAG's narrower null, which the posterior scale factor $\kappa$ corrects (Section~\ref{sec:ablation_study}).}
        
        \label{fig:mean_correlation}
    \end{subfigure}
    \label{fig:impact_kappa_and_mean_correlation}
\vspace{-0.5em}
\end{figure}

The central assumption underlying \attackName{} is that posterior samples from the SWAG approximation can substitute for independently trained reference models.
Since the attack accesses the per-target null $p(S \mid \mathcal{H}_0(\mathbf{x}))$ only through its moments $\mu(\mathbf{x}), \sigma(\mathbf{x})$, the fidelity of the approximation depends on recovering the same per-target score statistics, not on matching the weight-space distribution of the references. To isolate the source of gains, we compare LiRA and \attackName{} at matched ensemble size $K{=}N{=}4$ (Table~\ref{tab:k_matched_ablation}). Performance is comparable, confirming that individual SWAG samples match the quality to independently trained references. Increasing $K$ to 15, which is free for \attackName{} but requires 15 shadow models for LiRA, yields the improvements in Section~\ref{sec:results}, indicating that the advantage stems from cheaply averaging over a larger ensemble. Figure~\ref{fig:mean_correlation} corroborates this, as null means between SWAG and a 4-shadow ensemble agree strongly (Spearman $\rho = 0.926$), and the standardized scores $z(\mathbf{x})$ correlate well ($\rho = 0.771$).

This agreement holds because LoRA fine-tuning operates in a low-dimensional subspace atop a shared frozen backbone, so independently trained adapters remain within a comparable region of the loss landscape. SWAG models the SGD iterates within this region as a Gaussian posterior around the basin's mode~\citep{maddox2019simple}, capturing the score-level variability needed for calibration. The approximation is expected to degrade when independently trained references would genuinely separate, as our cross-adapter ablation (Table~\ref{tab:ablate_peft_qwen2vl}) confirms.

\subsection{Cross-adapter Transferability}
\label{sec:cross_transfer}
We relax the assumption that reference and target models share the same low-rank adaptation method. Beyond LoRA~\citep{hu2022lora}, we evaluate two PEFT methods: VeRA~\citep{kopiczko2024vera}, which shares frozen random low-rank matrices across layers and learns only lightweight scaling vectors, and LoRA-XS~\citep{balazy2024lora}, which inserts a small trainable matrix between frozen SVD-derived projections. Together they span a range of trainable-parameter budgets, probing robustness to adapter mismatch. Table~\ref{tab:ablate_peft_qwen2vl} compares \attackName{} with LiRA and RMIA. As expected, all reference-based attacks perform best when target and reference adapters match (diagonal entries). \attackName{} is more robust to moderate mismatch (e.g., LoRA$\leftrightarrow$VeRA): with a VeRA target, it achieves 27.33\% TPR at 3\% FPR on $\mathrm{D}_{\mathrm{real}}$, versus 20.67\% (LiRA) and 18.67\% (RMIA), despite using only $N{=}1$ reference model versus $N{=}4$ for the baselines. All reference-based attacks struggle, however, when the target uses extremely low-rank methods such as LoRA-XS.
\begin{table*}[t]
\centering

\begin{minipage}[t]{0.45\textwidth}
\vspace{0pt}
\centering
\small
\resizebox{\linewidth}{!}{%
\begin{tabular}{lccc}
\toprule 
\multirow{2}{*}{\textbf{Method}} & \multicolumn{2}{c}{\textbf{TPR@FPR}} & \multirow{2}{*}{\textbf{AUROC}} \\
\cmidrule(lr){2-3}
& \textbf{3\%} & \textbf{5\%} & \\
\midrule
LiRA ($N{=}4$)                & 22.17 & 30.50 & \textbf{79.42} \\
\attackName{} ($K{=}4$)$^{\dagger}$ & $27.18{\pm1.85}$ & $33.34{\pm1.63}$ & $77.09{\pm0.30}$ \\
\attackName{} ($K{=}15$)      & \textbf{31.67} & \textbf{34.33} & {77.92} \\
\bottomrule
\end{tabular}%
}
\captionof{table}{\textbf{$K$-matched Ablation} on Qwen2-VL. 
At $K{=}N{=}4$, \attackName{} matches LiRA; improves at $K{=}15$. ${\dagger}$: mean\,$\pm$\,std over 10 random $4$-sample subsets of $K{=}15$ posterior draws. Results reported on a validation set.}

\label{tab:k_matched_ablation}
\end{minipage}%
\hfill
\begin{minipage}[t]{0.54\textwidth}
\vspace{0pt}
\centering
\small
\resizebox{\linewidth}{!}{%
\begin{tabular}{llcccccc}
\toprule
\multirow{3}{*}{\textbf{Model}} 
& \multirow{3}{*}{\textbf{Method}} 
& \multicolumn{3}{c}{$\mathrm{D}_{\mathrm{real}}$} 
& \multicolumn{3}{c}{$\mathrm{D}_{\textcolor{red}{\mathrm{synth}}}$} \\
\cmidrule(lr){3-5} \cmidrule(lr){6-8}
& 
& \multicolumn{2}{c}{\textbf{TPR@FPR}} 
& \multirow{2}{*}{\textbf{AUC}} 
& \multicolumn{2}{c}{\textbf{TPR@FPR}} 
& \multirow{2}{*}{\textbf{AUC}} \\
\cmidrule(lr){3-4} \cmidrule(lr){6-7}
& 
& \textbf{3\%} & \textbf{5\%} 
& 
& \textbf{3\%} & \textbf{5\%} 
& \\
\midrule
\multirow{2}{*}{Idefics2}
& MC Dropout & 32.33 & 37.00 & 80.53 & 22.33 & 34.67 & 81.24 \\
& SWAG       & \textbf{38.00} & \textbf{44.67} & \textbf{82.42} & \textbf{31.67} & \textbf{38.00} & \textbf{83.46} \\
\midrule
\multirow{2}{*}{Qwen2-VL}
& MC Dropout & \textbf{42.00} & \textbf{51.67} & 81.53 & 13.00 & 23.00 & 70.36 \\
& SWAG       & 38.33 & 45.67 & \textbf{81.73} & \textbf{16.00} & \textbf{25.00} & \textbf{71.46} \\
\midrule
\multirow{2}{*}{DeepSeek-VL2}
& MC Dropout & 6.33 & 11.67 & 64.31 & 21.00 & 34.00 & {69.63} \\
& SWAG       & \textbf{14.00} & \textbf{17.33} & \textbf{67.04} & \textbf{24.33} & \textbf{34.67} & \textbf{81.64} \\
\bottomrule
\end{tabular}%
}
\captionof{table}{\textbf{Comparison of MC Dropout and SWAG.} Results on real and synthetic data for three VLMs.}
%TPR at 3\% and 5\% FPR and AUC, on real and synthetic subsets for Idefics2, Qwen2-VL, and DeepSeek-VL2.}
\label{tab:mc_dropout_swag_comparison}
\end{minipage}

\end{table*}

\begin{table*}[t]
\centering
\small
\resizebox{\textwidth}{!}{
\begin{tabular}{llcccccccccccc}
\toprule
&
& \multicolumn{6}{c}{$\mathrm{D}_{\mathrm{real}}$} 
& \multicolumn{6}{c}{$\mathrm{D}_{\textcolor{red}{\mathrm{synth}}}$} \\
\cmidrule(lr){3-8} \cmidrule(lr){9-14}
& 
& \multicolumn{2}{c}{\textbf{LoRA}} 
& \multicolumn{2}{c}{\textbf{VeRA}} 
& \multicolumn{2}{c}{\textbf{LoRA-XS}} 
& \multicolumn{2}{c}{\textbf{LoRA}} 
& \multicolumn{2}{c}{\textbf{VeRA}} 
& \multicolumn{2}{c}{\textbf{LoRA-XS}} \\
\cmidrule(lr){3-4} \cmidrule(lr){5-6} \cmidrule(lr){7-8}
\cmidrule(lr){9-10} \cmidrule(lr){11-12} \cmidrule(lr){13-14}
 & \textbf{Method}
& \textbf{3\%} & \textbf{5\%}
& \textbf{3\%} & \textbf{5\%}
& \textbf{3\%} & \textbf{5\%}
& \textbf{3\%} & \textbf{5\%}
& \textbf{3\%} & \textbf{5\%}
& \textbf{3\%} & \textbf{5\%} \\
\midrule
& Attack-P 
& 13.67 & 18.33 & 4.67 & 10.00 & 0.00 & 0.00
& 9.00 & 11.33 & 6.00 & 9.33 & 5.33 & 8.33 \\
& Q-MIA 
& 16.00 & 21.33 & 7.67 & 12.33 & 0.67 & 0.67
& 10.00 & 14.33 & 7.00 & 11.33 & \textbf{6.00} & \textbf{11.00} \\
\midrule
\multirow{3}{*}{\textbf{LoRA}}
& LiRA($N=4$)
& \textbf{40.67} & \textbf{47.33} & 20.67 & \textbf{32.67} & 3.67 & 6.00
& 8.00 & 8.33 & 8.00 & 10.33 & 2.33 & 3.67 \\
& RMIA($N=4$)
& 36.00 & 36.00 & 18.67 & 25.67 & 2.00 & 2.67
& 9.33 & 13.67 & \textbf{18.67} & 23.00 & 3.33 & 6.33 \\
& \attackName($N=1$)
& 38.33 & 45.67 & \textbf{27.33} & 30.67 & \textbf{5.67} & \textbf{6.67}
& \textbf{16.00} & \textbf{25.00} & 10.00 & \textbf{24.67} & 4.33 & 6.33 \\
\midrule
\multirow{3}{*}{\textbf{VeRA}}
& LiRA($N=4$)
& 34.33 & 42.67 & \textbf{33.67} & 38.33 & 1.00 & 3.67
& 7.67 & 14.00 & 14.67 & 20.60 & 3.67 & 5.00 \\
& RMIA($N=4$)
& 38.67 & 44.33 & 29.33 & 40.33 & 0.67 & 2.67
& 8.67 & 16.00 & 22.00 & 26.67 & 0.67 & 6.33 \\					
& \attackName($N=1$)
& \textbf{40.33} & \textbf{45.67} & 30.33 & \textbf{42.00}	& \textbf{6.67} & \textbf{8.00}
& \textbf{14.33} & \textbf{16.67} & \textbf{24.67} & \textbf{30.33} & \textbf{5.33} & \textbf{7.67} \\
% \midrule
% \multirow{3}{*}{\textbf{LoRA-XS}}
% & LiRA($N=4$)
% & -- & -- & -- & -- & -- & --
% & -- & -- & -- & -- & -- & -- \\
% & RMIA($N=4$)
% & -- & -- & -- & -- & -- & --
% & -- & -- & -- & -- & -- & -- \\
% & \attackName($N=1$)
% & -- & -- & -- & -- & -- & --
% & -- & -- & -- & -- & -- & -- \\
\bottomrule
\end{tabular}
}
\caption{\textbf{Cross-adapter transferability on Qwen2-VL for DocVQA}. Rows: adapter used for reference models; columns: target adapter. Attack-P/QMIA are reference-free baselines. Trainable parameters: LoRA (5M), VeRA (720K), LoRA-XS (10K). Best results per cross-transfer are in \textbf{bold}.}
\label{tab:ablate_peft_qwen2vl}
\end{table*}

\subsection{Ablation Study}
\label{sec:ablation_study}
\textbf{SWAG vs.\ Alternative Bayesian Inference.} \attackName{} relies on a posterior over fine-tuned weights, instantiated with SWAG~\citep{maddox2019simple}, which fits a low-rank plus diagonal Gaussian along the SGD trajectory. Two natural alternatives are Laplace approximation~\citep{yang2024bayesian}, which does not extend to open-ended generative tasks, and MC Dropout~\citep{pmlr-v48-gal16}, which reuses fine-tuning dropout as its posterior surrogate. Table~\ref{tab:mc_dropout_swag_comparison} compares SWAG with MC Dropout across three target VLMs. SWAG dominates on Idefics2 and DeepSeek-VL2 across all metrics on both subsets, which we attribute to its trajectory-based posterior capturing weight correlations informative for distinguishing members from non-members.
% \begin{wrapfigure}{r}{0.52\textwidth}
%     \centering
%     \includegraphics[width=0.5\textwidth]{graphics/impact_K_Qwen2VL.pdf}
%     \caption{Your figure caption}
%     \label{fig:ablate_k_qwen2vl}
% \end{wrapfigure}

% \begin{figure}[htbp]
%     \centering
%     \includegraphics[width=\linewidth]{graphics/impact_K_Qwen2VL.pdf}
%     \caption{Your figure caption}
%     \label{fig:ablate_k_qwen2vl}
% \end{figure}

\begin{figure}[htbp]
    \centering
    \begin{subfigure}{0.45
    \textwidth}
        \centering
        \includegraphics[width=\linewidth]{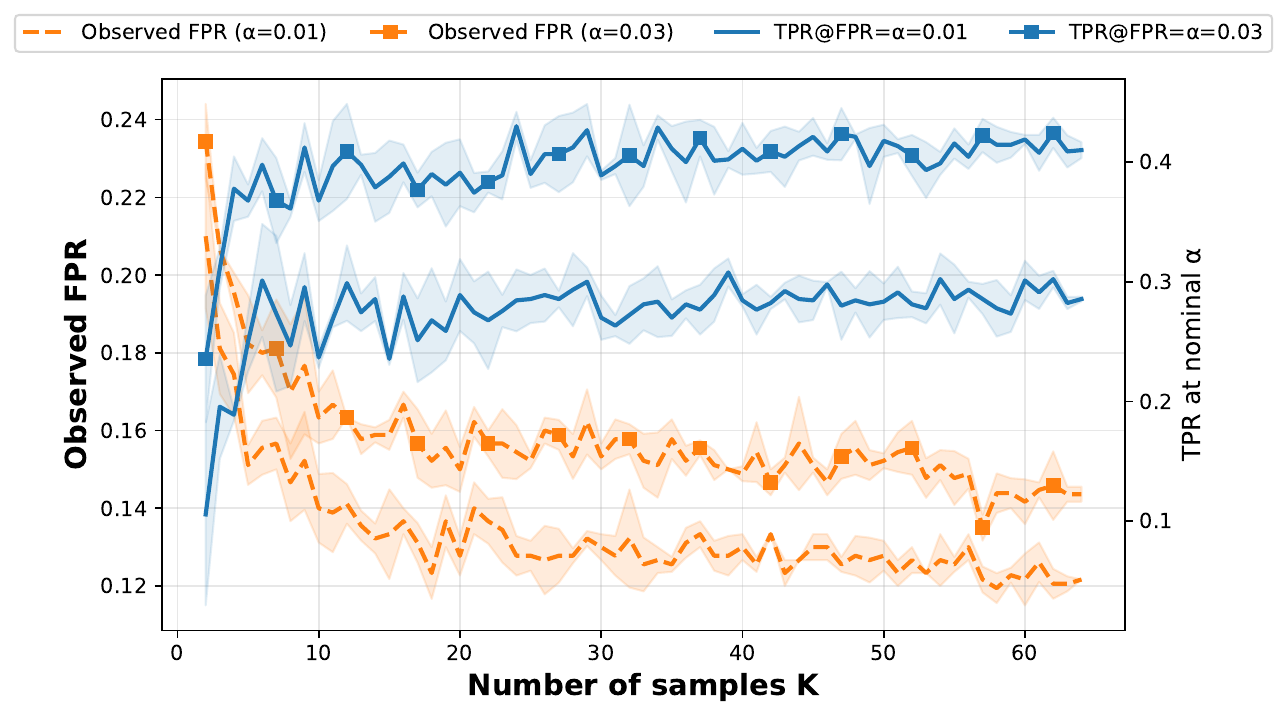}
        \caption{\textbf{Effect of ensemble size $K$ on Qwen2-VL.} TPR at nominal FPR levels $\alpha \in \{0.01, 0.03\}$ (right axis, solid) and the corresponding observed FPR (left axis, dashed), averaged over 5 runs.}
        \label{fig:ablate_num_samples_qwen2vl}
    \end{subfigure}
    \hfill
    \begin{subfigure}{0.54\textwidth}
        \centering
        \includegraphics[width=\linewidth]{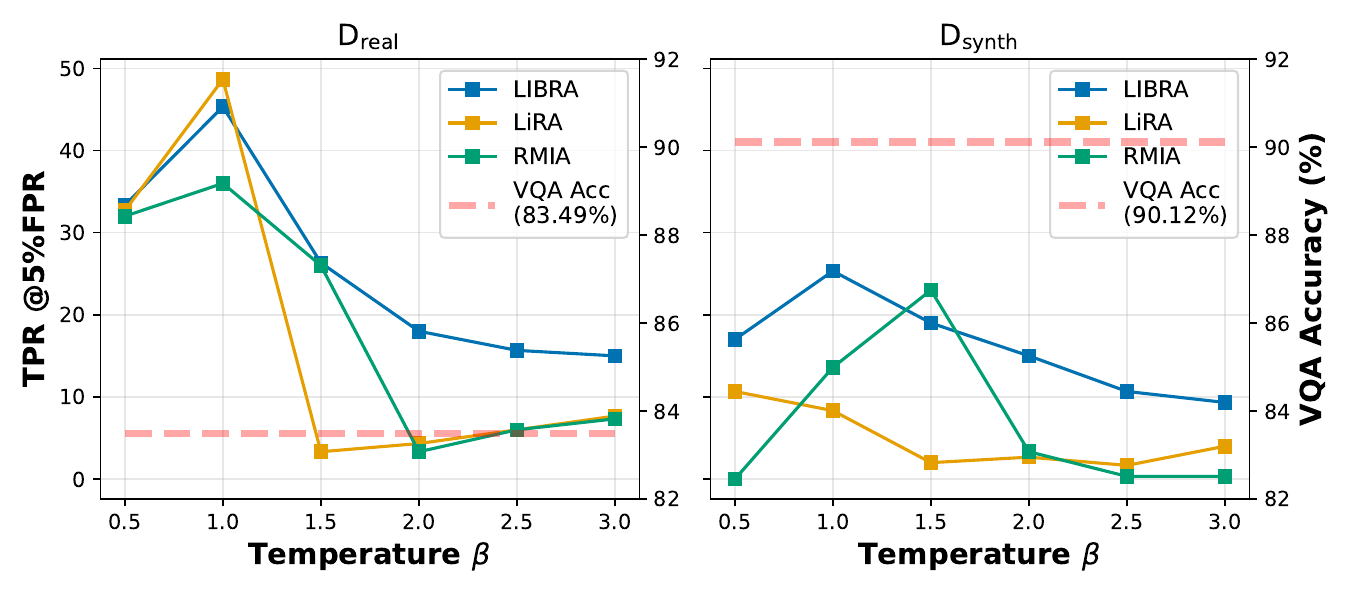}
        \caption{\textbf{Effect of inference temperature $\beta$ on Qwen2-VL.} Dashed line shows VQA accuracy (right axis), which is preserved under greedy decoding. \attackName{} degrades more gracefully than LiRA and RMIA as $\beta$ grows.}
        \label{fig:ablate_temperature_qwen2vl}
    \end{subfigure}
    \label{fig:ablate_qwen2vl}
\vspace{-0.5em}
\end{figure}

\textbf{Effect of ensemble size $K$.} Figure~\ref{fig:ablate_num_samples_qwen2vl} reports \attackName{}'s performance under \protocoltwo{} (per-sample threshold) as we vary $K$. TPR at both nominal FPR levels saturates around $K{=}15$, which we adopt as the default throughout the paper. The observed FPR decreases as $K$ grows, indicating progressively better calibration of the null with more samples, although a gap to the nominal level remains.

\noindent\textbf{Effect of scale factor $\kappa$.} Since the SWAG null is correctly centered but too narrow, the over-rejection under the per-sample protocol is a second-moment effect. Sweeping $\kappa$, which scales the posterior covariance (Figure~\ref{fig:ablate_kappa}), confirms this:
dispersion $\sigma_{\text{SWAG}}/\sigma_{\text{shadow}}$ increases monotonically, observed FPR drops from $12.67\%$ at $\kappa{=}2$ to $9\%$ at $\kappa{=}8$, while TPR and AUC remain flat or improve. A residual gap persists at dispersion parity ($\kappa{=}10$), indicating that remaining miscalibration arises from the Gaussian tail approximation in Eq.~\ref{eq:z_score}, which is a limitation shared by all LiRA-style attacks~\citep{carlini2022membership}.

\textbf{Effect of inference temperature $\beta$.} Figure~\ref{fig:ablate_temperature_qwen2vl} examines a simple defense: applying a temperature $\beta$ to the target model's output distribution at inference. Higher $\beta$ flattens the next-token distribution, smoothing the likelihood gap between members and non-members while leaving VQA accuracy unchanged under greedy decoding, the \emph{standard} inference setup for VQA tasks. All reference-based attacks weaken as $\beta$ grows, but \attackName{} degrades more gracefully and maintains a consistent margin on $\mathrm{D}_{\mathrm{synth}}$. 
We attribute this to \attackName{}'s posterior-based scoring, which aggregates evidence across weight samples rather than relying on a single likelihood that temperature directly attenuates.

% \section{Defenses.}
% \label{sec:defenses}
% We study different factors affecting the privacy risks of fine-tuned models including target model sizes and training epochs and how these impact different MIA methods

\section{Limitations}
\label{sec:limitations}
Our approach inherits several limitations of reference-based MIAs. First, attack performance relies on how faithfully the posterior captures the variability of independently fine-tuned models, which may degrade under adapter mismatch or non-standard tuning (Table~\ref{tab:ablate_peft_qwen2vl}). Second, the Gaussian relaxation of the per-target null limits low-FPR tail calibration, a constraint shared by all LiRA-style attacks (Figure~\ref{fig:ablate_kappa}). Non-parametric or heavier-tailed null models are a direction for future work. Third, inference cost scales linearly with $K$, since each target is scored under all $K$ virtual references, though total training cost remains substantially lower than reference-model ensembles (Table~\ref{tab:wallclock_time}). Finally, while we focus on LoRA and SWAG, \attackName{} applies to any trainable parameter subset admitting optimization snapshots and is compatible with alternative posteriors such as deep ensembles~\citep{onal2024gaussian}.

\section{Conclusion}
\label{sec:conclusion}
We introduced \attackName{}, a scalable membership inference attack for parameter-efficient fine-tuned VLMs that replaces costly ensembles of independently trained reference models with a Bayesian posterior over LoRA parameters. Our approach preserves the statistical advantages of reference-model calibration while dramatically reducing the training cost associated with these attacks through virtual reference models sampled from a single SWAG approximation. Experiments on DocVQA and Medical VQA across three distinct VLMs show that \attackName{} consistently matches or surpasses existing reference-based attacks despite requiring only a single fine-tuning run. These results demonstrate that reliable privacy auditing of fine-tuned VLMs is feasible without training large shadow-model ensembles, suggesting that computational cost alone should not be considered a practical barrier against rigorous membership inference and privacy auditing in sensitive multi-modal applications.

\clearpage
\newpage

\acksection{
    This work has been funded by the European Union's Horizon Europe programme through the European Lighthouse on Safe and Secure AI (ELSA) under grant agreement No 101070617 and the European Large Open Multi-Modal Foundation Models For Robust Generalization On Arbitrary Data Streams (ELLIOT) under grant agreement No 101214398. Views and opinions expressed are however those of the authors only and do not necessarily reflect those of the European Union or European Commission.
    Khanh Nguyen, Andrey Barsky, and Dimosthenis Karatzas have been supported by the Consolidated Research Group 2021 SGR 01559 from the Research and University Department of the Catalan Government, and by project PID2023-146426NB-100 funded by MCIU/AEI/10.13039/501100011033 and FSE+.
    Lastly, this work has also been partially supported by project FISA-2023-00128 funded by the MUR program ``Fondo italiano per le scienze applicate". 
}

\bibliography{iclr2026_conference}
\bibliographystyle{iclr2026_conference}
\medskip

{
%%%%%%%%%%%%%%%%%%%%%%%%%%%%%%%%%%%%%%%%%%%%%%%%%%%%%%%%%%%%

\appendix

\clearpage

\section{Preliminaries}
\label{app:preliminaries}
LoRA~\cite{hu2022lora} freezes the pretrained weights $W_0 \in \mathbb{R}^{d \times k}$ and learns low-rank updates:
\begin{equation}
    W = W_0 + \Delta W = W_0 + BA,
\end{equation}
where $B \in \mathbb{R}^{d \times r}$, $A \in \mathbb{R}^{r \times k}$, and $r \ll \min(d, k)$. Only $A$ and $B$ are optimized, substantially reducing the number of trainable parameters while maintaining strong downstream performance. 
Due to its efficiency and effectiveness, LoRA and related variants are widely used to adapt VLMs on proprietary data before deployment or release~\cite{kopiczko2024vera, balazy2024lora, labrak2024biomistral, chen2025svrag}.

\section{Additional Dataset Details}
\subsection{Datasets}
\label{app:dataset}
\noindent\textbf{PFL-DocVQA}\citep{tito2024privacy} is a large-scale dataset of real business invoices that frequently contain privacy-sensitive information such as payment amounts, tax identification numbers, and bank account details. The dataset is specifically designed for DocVQA and is accompanied by a MIA variant, in which the objective is to infer whether a given invoice provider's documents were included in the training set, based on a held-out set of their invoices. Following the original construction protocol, we design a synthetic pipeline that mimics both the visual style of the invoices and the corresponding question-answer pairs. Since the QA task is extractive, where answers can be located directly within the document image, we use this dataset for the \emph{multi-sample} and \emph{image-only} MIA settings. 

\noindent\textbf{SLAKE}\citep{liu2021slake} is a bilingual Medical VQA dataset of radiology images (CT, MRI, and X-ray) covering five body regions, paired with question–answer pairs annotated by experienced physicians. We use SLAKE in the \emph{sample-level} MIA setting, as its high-quality textual annotations constitute proprietary assets that warrant privacy protection. No synthetic data is available for this dataset.

% \input{figures/questions_per_document}
% Figure \ref{fig:questions_per_document} illustrates the distribution of the number of questions per document. Notably, the distribution is relatively skewed. We refer readers to the original papers for further details on dataset construction and statistics.

\subsection{Document Synthesis Pipeline}
\label{app:synthesis}

For generation of synthetic invoices, we aim to mimic a large manifold within PFL-DocVQA that comprises advertising invoices for US political campaigns. Many of these providers use identical layout templates that differ only in specific names, numbers, addresses, logos and line item totals; this observation inspired our approach to create semi-realistic new invoices that are guaranteed to contain nothing that can be memorized from the pre-training data.

\begin{figure}[htbp]
    \centering
    \includegraphics[width=\linewidth]{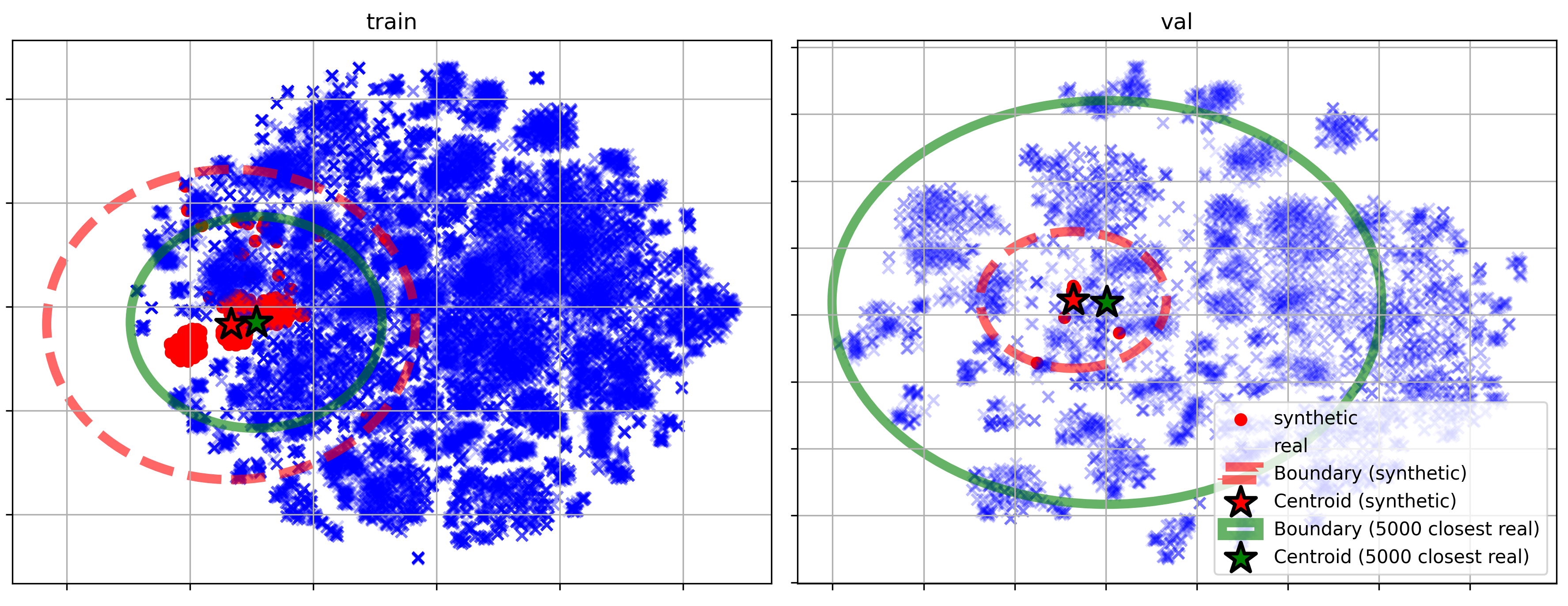}
    \caption{\textbf{Centroid-based selection of real documents from PFL-DocVQA.} t-SNE projection of pre-finetuning CLS-token embeddings from Idefics2. Real documents (\textcolor{blue}{blue} $\times$) span a much broader region than synthetic ones (\textcolor{red}{red} $\bullet$), so we form $\mathrm{D}_{\mathrm{real}}$ from the 5000 real documents closest in Euclidean distance to the synthetic centroid (\textcolor{red}{red} $\star$); the resulting cluster centroid (\textcolor{green}{green} $\star$) nearly coincides with the synthetic one. This ensures that pre-finetuning representational gaps cannot be exploited as a membership shortcut. Splits are constructed independently for each target VLM.}
    \label{fig:document_embedding_docvqa}
\end{figure}

Initially, we randomly choose a base invoice template (as a spreadsheet) from a selection containing various layouts and table formats. This template is populated with a selection of invoice fields using similar verbiage as in PFL-DocVQA ("Advertiser", "Contract number", "Invoice Period" etc.). For each field, we randomly generate plausible values within each provider that observe the contextual and syntactic constraints required, e.g.: each billing address is randomly drawn from a list of US streets in real cities; provider contact details follow the NANP scheme for fictional US phone numbers; personal names of advertisers and salespeople are sampled from statistical US naming data; line items for ad slots are randomly generated with plausible durations and date/time intervals with plausible per-minute billing rates that sum to the correct totals. 
Names and logos of providers are drawn from a freely available dataset of placeholder graphics (\url{https://uilogos.co/}). 

Since the forward generation process is imperative and fully controlled, we also exploit the ability to generate annotated question-answer pairs for each new document (also of the same extractive format as in PFL-DocVQA) that can be used for fine-tuning. 

\noindent \textbf{Selection of real documents.} Figure~\ref{fig:document_embedding_docvqa} illustrates the similarity of the synthetic datapoints to the broader PFL-DocVQA distribution. As observations, real documents span a much broader embedding region than synthetic ones. Evaluating on the full real distribution risks artificially inflating MIA performance, since distributional outliers are easier to memorize and attack scores would partly reflect anomaly rather than membership signal~\citep{carlini2022membership, watson2022on}. To avoid this confound, we select $\mathrm{D}_{\mathrm{real}}$ as the real documents closest to the synthetic centroid in frozen, pre-fine-tuning embedding space, ensuring that the two evaluation subsets occupy a comparable region of the feature space. Crucially, this selection is performed before and independently of membership-label assignment: documents are first selected, then randomly split into balanced member and non-member sets. Results are reported separately for $\mathrm{D}_{\mathrm{real}}$ and $\mathrm{D}_{\mathrm{synth}}$ throughout. This protocol yields a more conservative evaluation that isolates the membership signal from distributional artifacts, at the cost of narrower coverage of real-world variability.

\begin{figure}[htbp]
    \centering

    \begin{subfigure}{0.24\textwidth}
        \centering
        \fbox{\includegraphics[width=\linewidth]{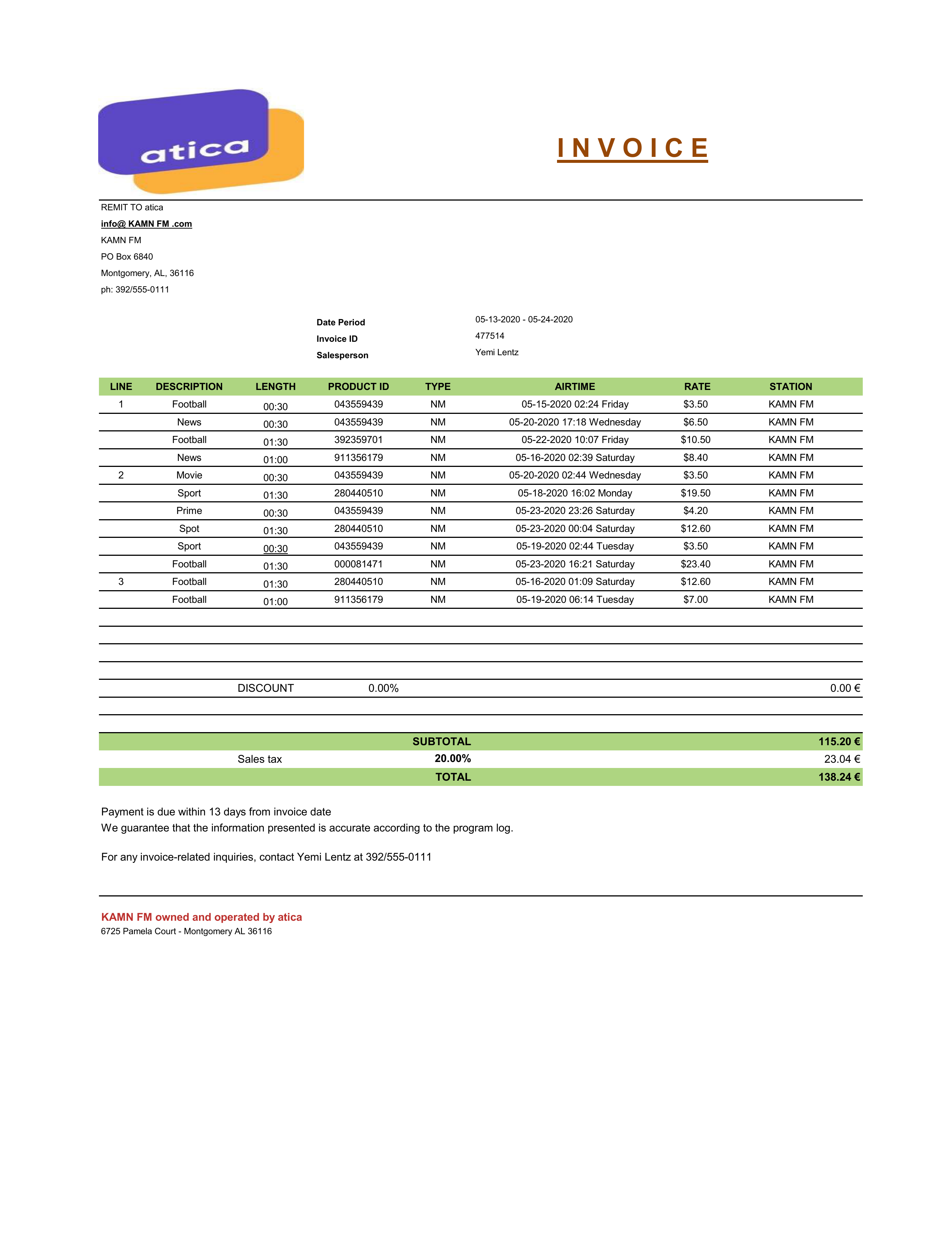}}
    \end{subfigure}
    \hfill
    \begin{subfigure}{0.24\textwidth}
        \centering
        \fbox{\includegraphics[width=\linewidth]{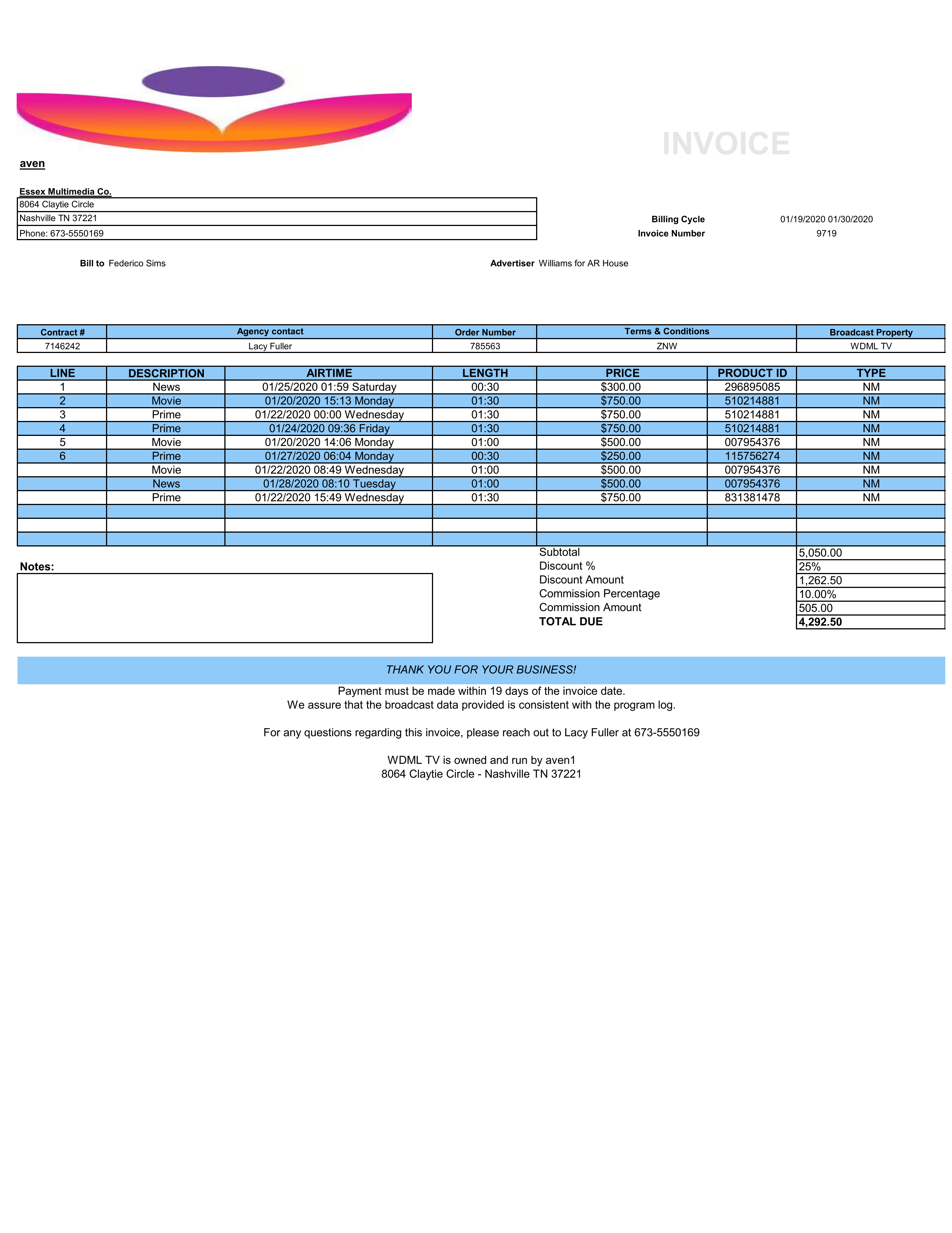}}
    \end{subfigure}
    \hfill
    \begin{subfigure}{0.24\textwidth}
        \centering
        \fbox{\includegraphics[width=\linewidth]{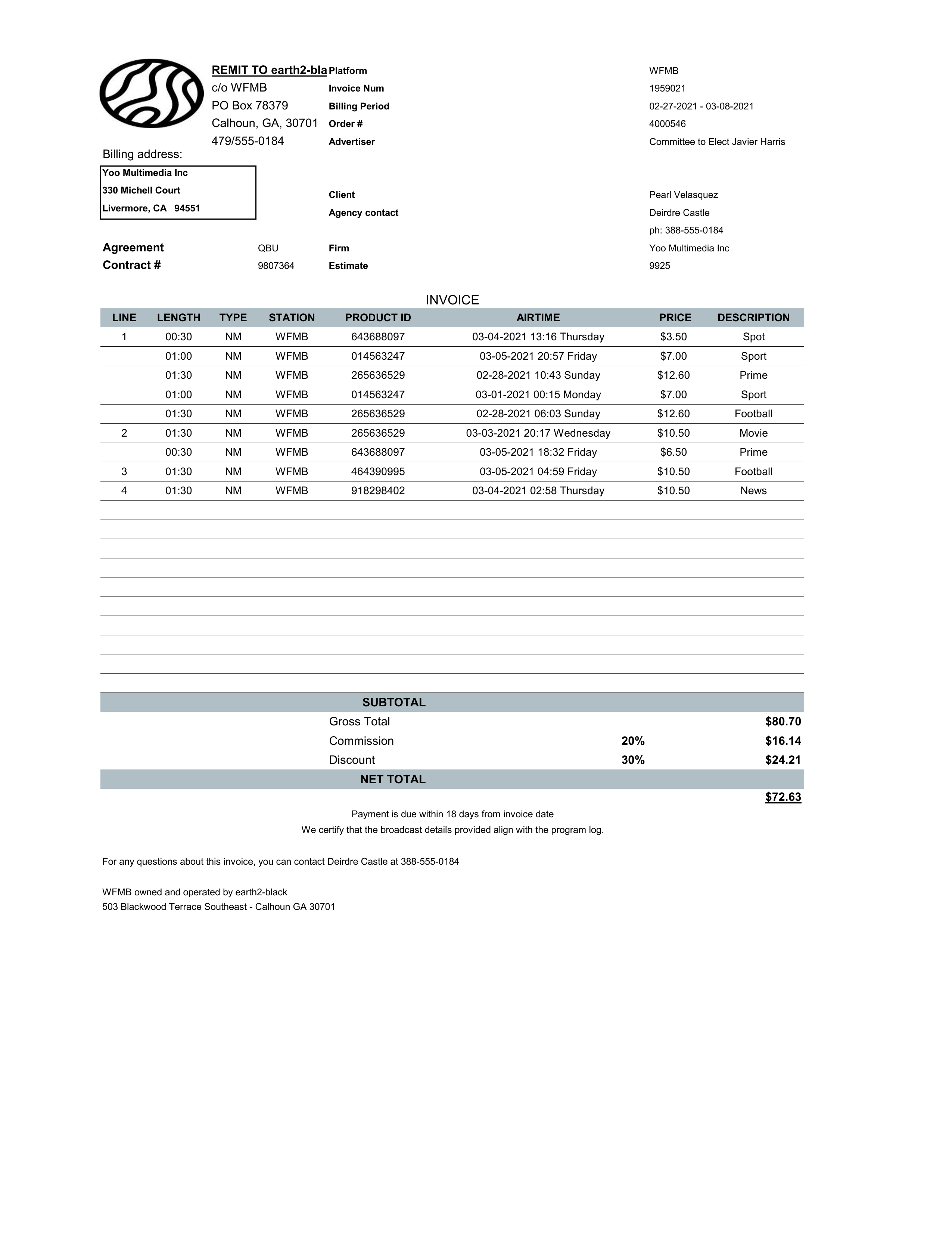}}
    \end{subfigure}
    \hfill
    \begin{subfigure}{0.24\textwidth}
        \centering
        \fbox{\includegraphics[width=\linewidth]{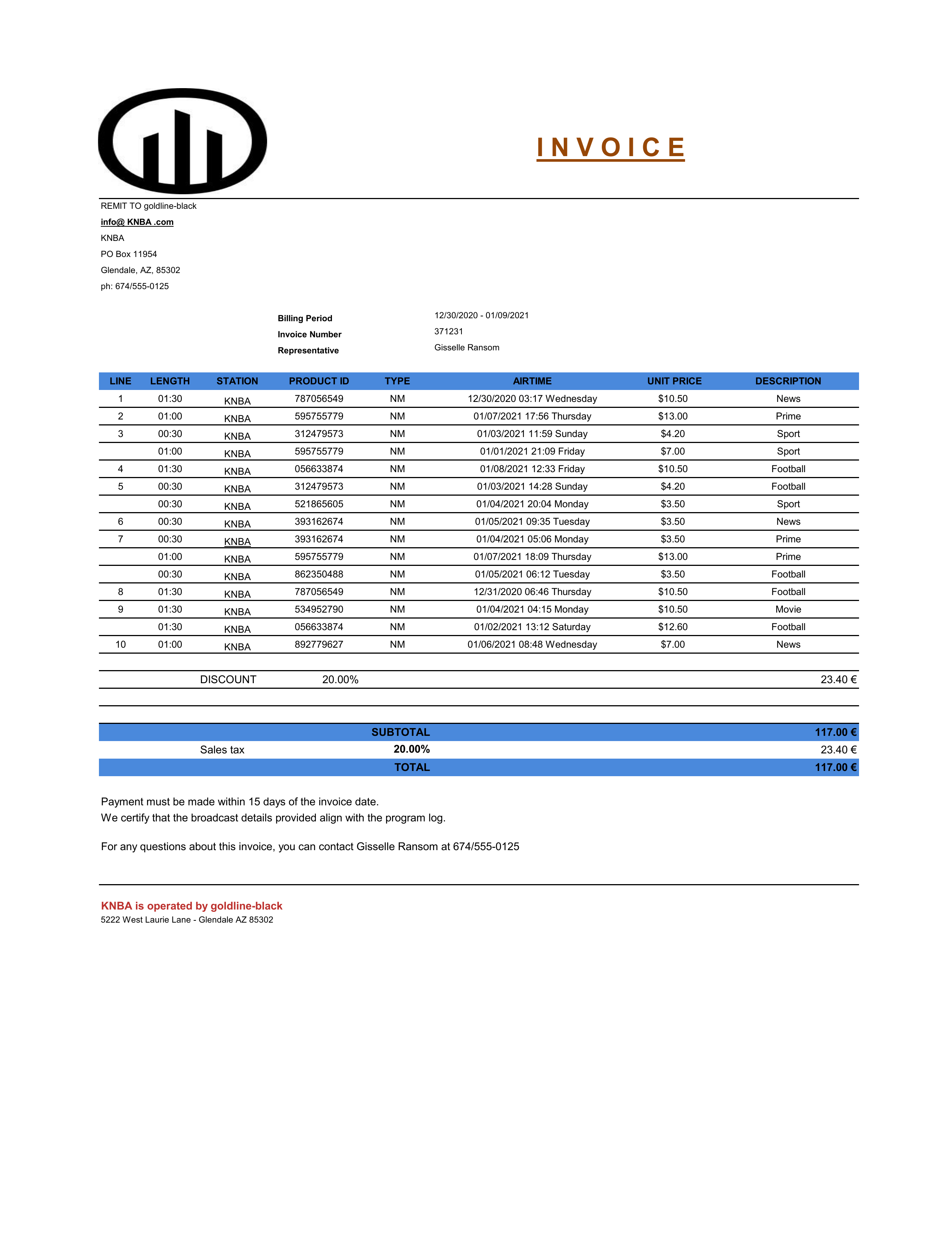}}
    \end{subfigure}

    \begin{subfigure}{0.24\textwidth}
        \centering
        \fbox{\includegraphics[width=\linewidth]{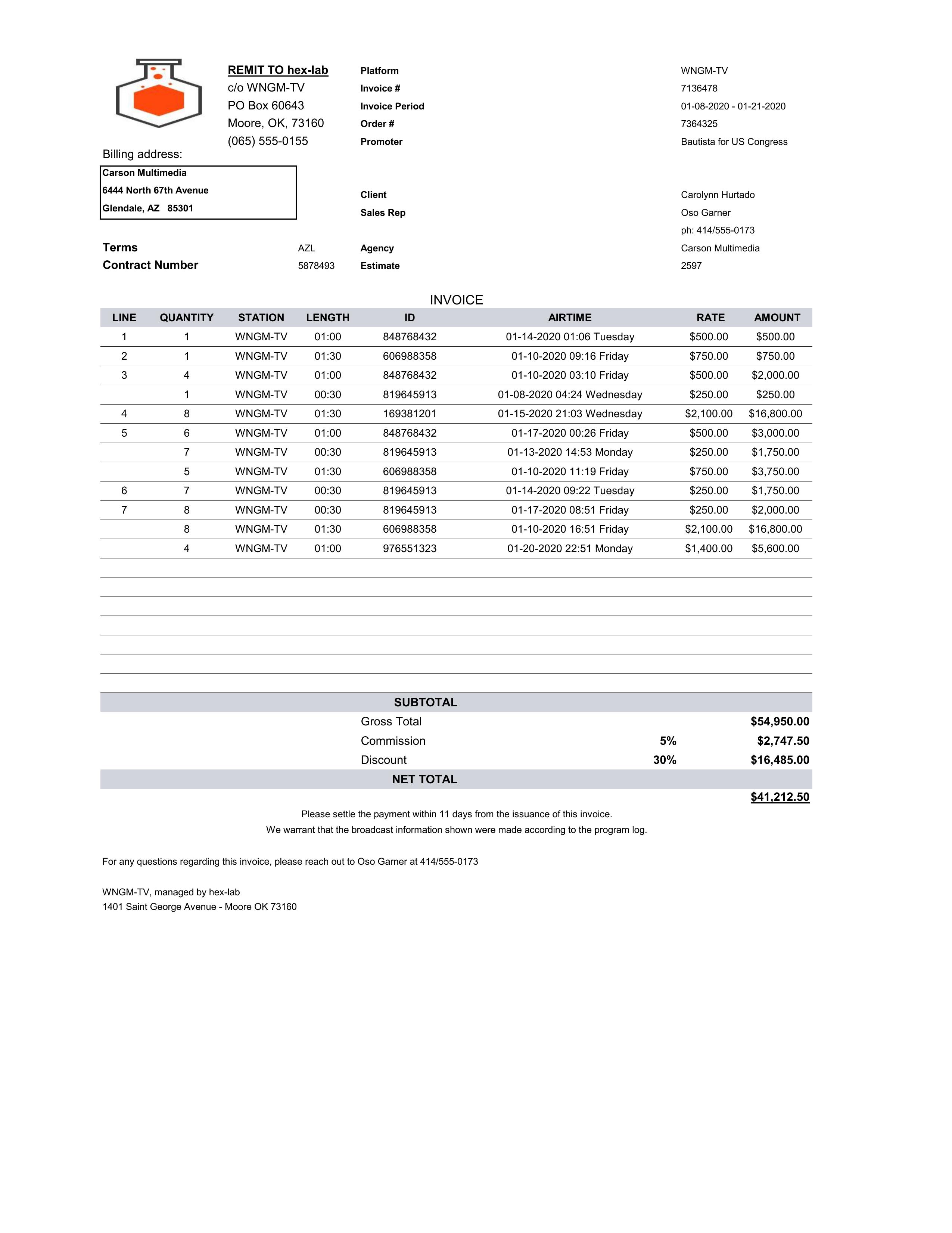}}
    \end{subfigure}
    \hfill
    \begin{subfigure}{0.24\textwidth}
        \centering
        \fbox{\includegraphics[width=\linewidth]{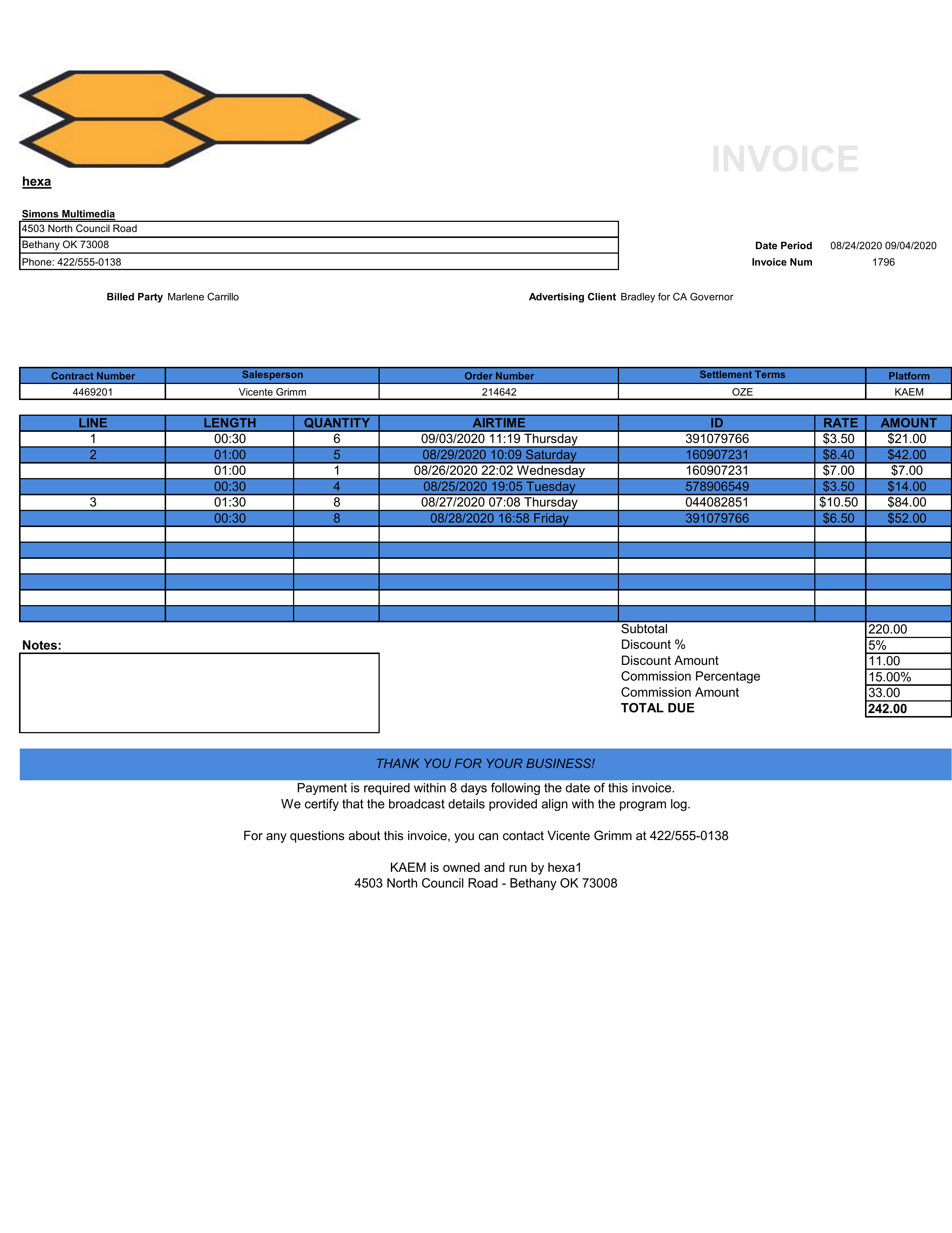}}
    \end{subfigure}
        \hfill
    \begin{subfigure}{0.24\textwidth}
        \centering
        \fbox{\includegraphics[width=\linewidth]{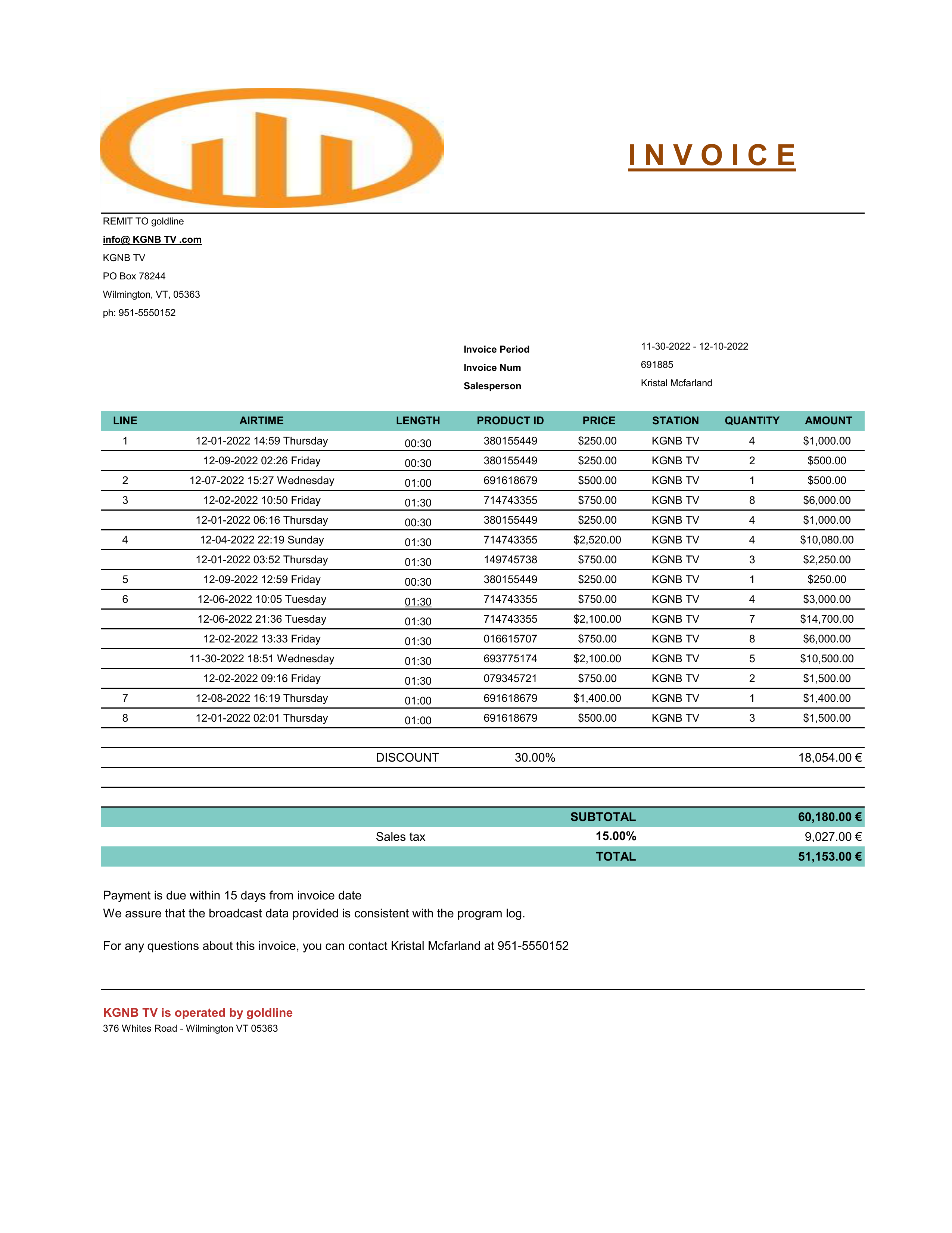}}
    \end{subfigure}
    \hfill
    \begin{subfigure}{0.24\textwidth}
        \centering
        \fbox{\includegraphics[width=\linewidth]{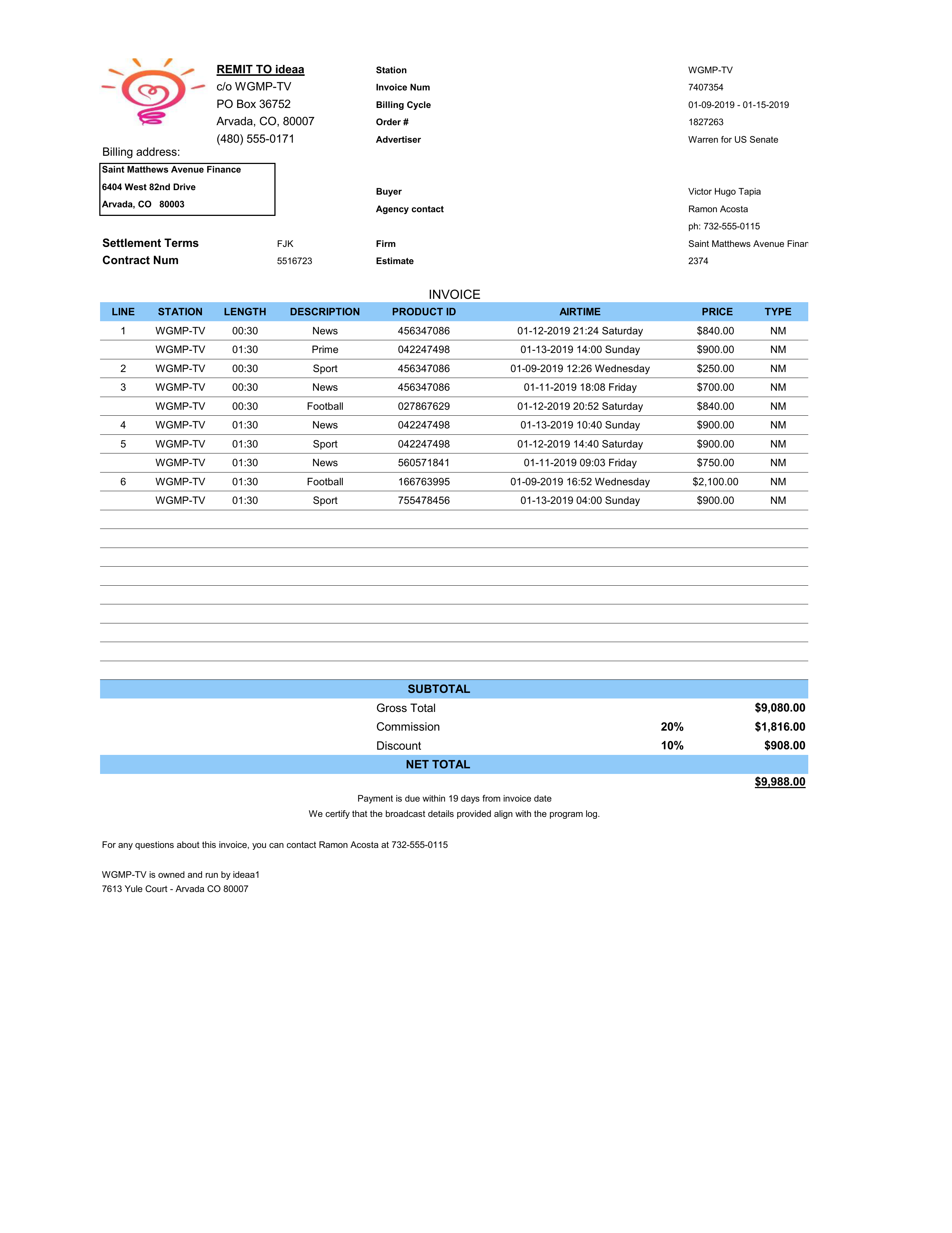}}
    \end{subfigure}
    \caption{\textbf{Synthetic documents for DocVQA} generated by our document synthesis pipeline.}
    \label{fig:docvqa_synthetic_documents}
\end{figure}
\begin{figure}[htbp]
    \centering
    \hfill
    \begin{subfigure}[t]{0.32\textwidth}
        \centering
        \fbox{\includegraphics[width=\linewidth]{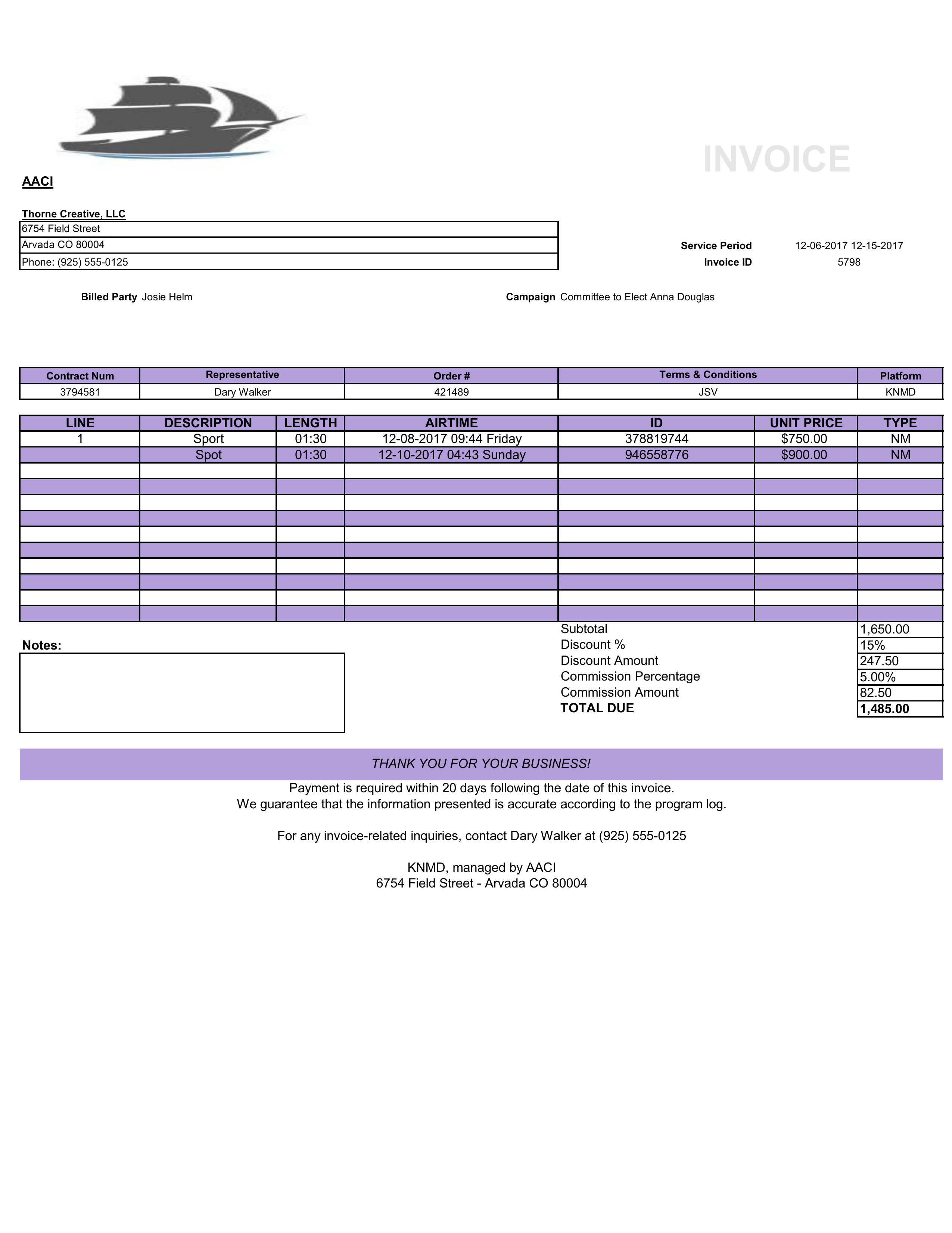}}
        \vspace{4pt}
        \begin{minipage}{\linewidth}
            \raggedright
            \footnotesize
            \textbf{Q:} What is the agency's name?\\
            \textbf{A:} Thorne Creative, LLC\\
            \textbf{Q:} What is the aggregate total?\\
            \textbf{A:} 1650.0
        \end{minipage}
    \end{subfigure}
    \hfill
    \begin{subfigure}[t]{0.32\textwidth}
        \centering
        \fbox{\includegraphics[width=\linewidth]{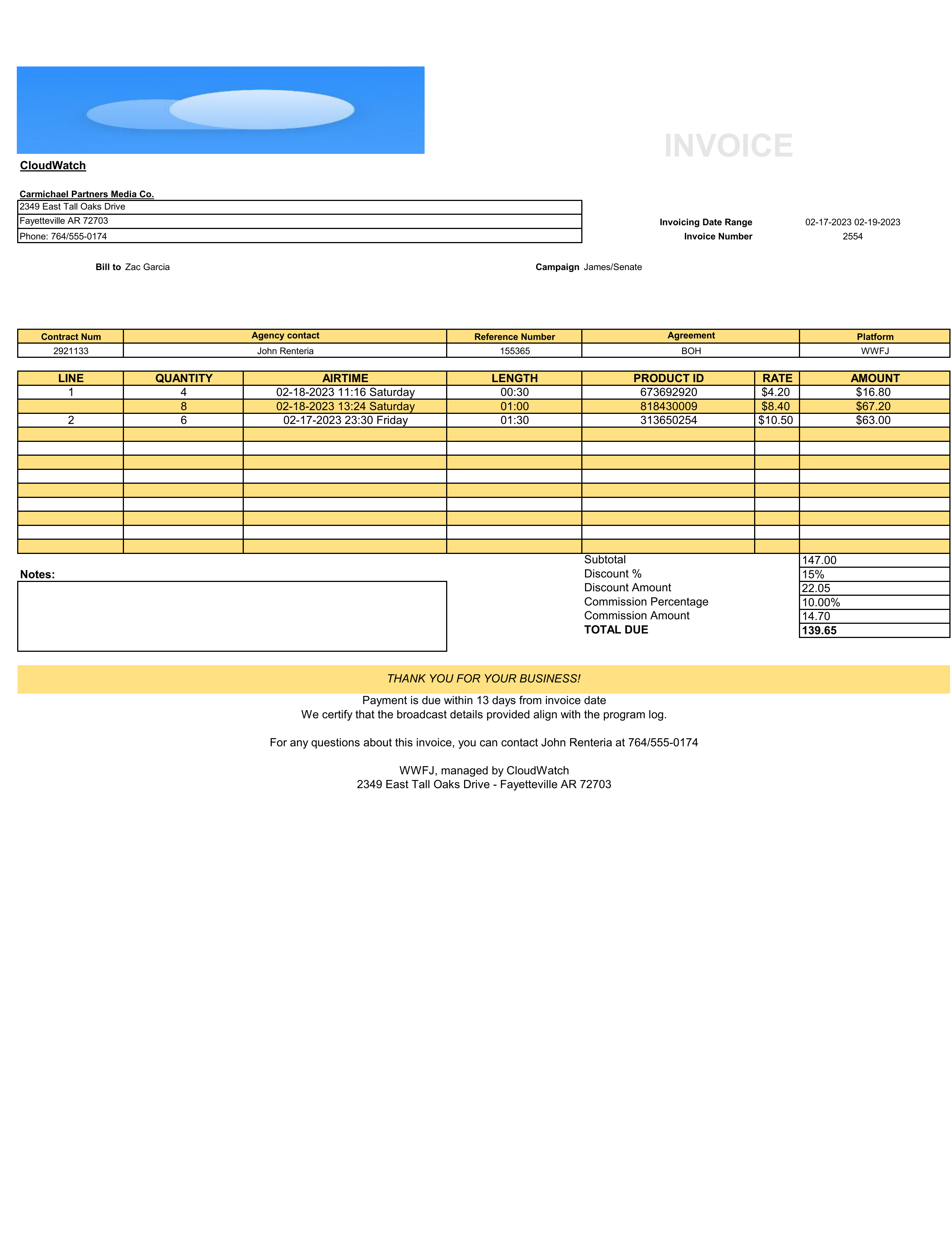}}
        \vspace{4pt}
        \begin{minipage}{\linewidth}
            \raggedright
            \footnotesize
            \textbf{Q:} What is the phone number?\\
            \textbf{A:} 764/555-0174\\
            \textbf{Q:} Can you provide information about the payment time?\\
            \textbf{A:} start: 02-17-2023, end: 02-19-2023
        \end{minipage}
    \end{subfigure}
    \hfill
    \begin{subfigure}[t]{0.32\textwidth}
        \centering
        \fbox{\includegraphics[width=\linewidth]{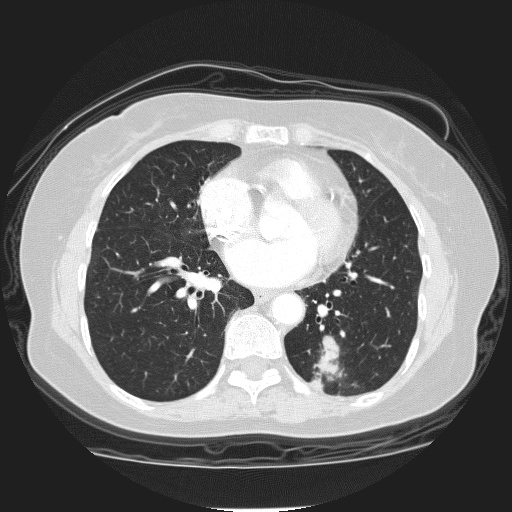}}
        \vspace{4pt}
        \begin{minipage}{\linewidth}
            \raggedright
            \footnotesize
            \textbf{Q:} What diseases are included in the picture?\\
            \textbf{A:} Lung Cancer\\
            \textbf{Q:} What is the main organ in the image?\\
            \textbf{A:} Lung, Heart
        \end{minipage}
    \end{subfigure}
    \hfill
    \caption{\textbf{VQA examples.} The first two (synthesized) examples are from Document VQA, and the third is from Medical VQA.}
    \label{fig:docvqa_synthetic_examples}
\end{figure}

\section{Further details on \attackName{}}

We detail the steps of our proposed attack \attackName{} in Algorithm~\ref{alg:algorithm}. Following the evaluation protocol of LiRA~\cite{carlini2022membership} and RMIA~\cite{zarifzadeh2023low}, we sweep a global threshold $\tau$ over the attach score $z(\mathbf{x})$ to trace the full ROC curve, rather than committing to a target-specific $\tau_\alpha(\mathbf{x})$.

\begin{algorithm}[t]
\caption{Low-rank Membership Inference via Bayesian Reference Approximation (\attackName)}
\label{alg:algorithm}
\begin{algorithmic}[1]
\Require Target VLM (frozen) with adapter $\boldsymbol{\theta}_0$, target instance $\mathbf{x}$, shadow dataset $\mathrm{D}_{\mathrm{shadow}}$, number of reference samples $K$, SWAG rank $M$.
\State $S(\mathbf{x}) \gets s(\mathbf{x}; \boldsymbol{\theta}_0)$
\State Fine-tune adapter on $\mathrm{D}_{\mathrm{shadow}}$ to obtain $\boldsymbol{\theta}^{\star}$
\State Collect SGD iterates $\{\boldsymbol{\theta}_s\}_{s=1}^{T}$ from $\boldsymbol{\theta}^{\star}$ with constant learning rate
\State Compute SWAG moments $\bar{\boldsymbol{\theta}}$, $\Sigma_{\mathrm{diag}}$, and deviation matrix $\hat{D} \in \mathbb{R}^{p \times M}$
\For{$k = 1$ \textbf{to} $K$}
    \State Sample $\mathbf{z}_1^{(k)} \sim \mathcal{N}(0, I_p)$, $\mathbf{z}_2^{(k)} \sim \mathcal{N}(0, I_M)$
    \State $\boldsymbol{\theta}^{(k)} \gets \bar{\boldsymbol{\theta}} + \tfrac{1}{\sqrt{2}}\,\Sigma_{\mathrm{diag}}^{1/2}\,\mathbf{z}_1^{(k)} + \tfrac{1}{\sqrt{2(M-1)}}\,\hat{D}\,\mathbf{z}_2^{(k)}$
    \State $S_k \gets s(\mathbf{x}; \boldsymbol{\theta}^{(k)})$ \Comment{reference score from $k$-th virtual model}
\EndFor
\State $\mu(\mathbf{x}) \gets \dfrac{1}{K} \sum_{k=1}^{K} S_k$
\State $\sigma(\mathbf{x}) \gets \sqrt{\dfrac{1}{K-1} \sum_{k=1}^{K} \bigl(S_k - \mu(\mathbf{x})\bigr)^2}$
\State \Return Standardized score $z(\mathbf{x}) \gets \dfrac{S(\mathbf{x}) - \mu(\mathbf{x})}{\sigma(\mathbf{x})}$
\end{algorithmic}
\end{algorithm}

\section{Implementation Details}
\label{app:impl_details}
All experiments are run on a single node with NVIDIA A40 GPUs with 45GB memory.

\subsection{Data splits.}
We assume a small fine-tuning regime and construct balanced training and test sets, $\mathrm{D}_{\mathrm{train}}$ and $\mathrm{D}_{\mathrm{test}}$. For \emph{multi-sample} MIA on DocVQA, each split contains 300 real and 300 synthetic documents, yielding disjoint evaluation subsets $\mathrm{D}_{\mathrm{real}}$ and $\mathrm{D}_{\mathrm{synth}}$ of 600 documents each. The 300 real documents are chosen as those closest to the centroid of the synthetic cluster in the target model's pre-finetuning embedding space, minimizing distribution shift between $\mathrm{D}_{\mathrm{real}}$ and $\mathrm{D}_{\mathrm{synth}}$ so that proximity in feature space cannot itself drive the MIA decision (Figure~\ref{fig:document_embedding_docvqa}). Splits are therefore target-model-specific. For \emph{image-only} MIA, where the membership unit is a single image $v$, we rephrase each original question with Qwen2.5-14B-Instruct conditioned on the document; since answers are grounded in the image, these rephrasings approximate the original training annotations. We subsample the rephrased pool to vary how often each target image is seen during fine-tuning. For \emph{sample-level} MIA on MedVQA, where membership is defined on a $(v, q, a)$ tuple, each split draws an equal number of samples from the original data, so an image may appear in both member and non-member sets paired with different QA pairs. The adversary is given a \emph{shadow} dataset of twice the size of $\mathrm{D}_{\mathrm{train}}$, balanced and disjoint from it; it serves as the population for QMIA's quantile regressor and as the source from which reference models are trained on random half-subsets. A separate balanced split of one quarter the size of $\mathrm{D}_{\mathrm{train}}$ is used as the validation set for QMIA and as the population set $\mathcal{Z}$ for RMIA.

\subsection{Target Model Training}
\label{app:target_model}
We fine-tune all target VLMs from publicly available HuggingFace checkpoints using LoRA~\citep{hu2022lora} (via the PEFT library~\citep{peft}), applied to the linear projections in the latter half of the LLM's transformer layers. We fix the LoRA rank to $r=4$ and dropout to $0.2$, as the low-data regime makes VLMs prone to overfitting. Training is performed with Adam for 3 epochs at an effective batch size of 16 via gradient accumulation. Per-model hyperparameters are reported in Table~\ref{tab:training_hps}. Unless otherwise noted, we report MIA results after the first epoch—the most challenging regime, in which each private sample is observed exactly once.

\begin{table*}[t]
\centering
\begin{minipage}[t]{0.48\textwidth}
\centering
\small
\resizebox{\linewidth}{!}{\begin{tabular}{llccc}
\toprule
\multirow{2}{*}{\textbf{Model}} & \multirow{2}{*}{\textbf{Method}} & \multicolumn{2}{c}{\textbf{PFL-DocVQA}} & \textbf{SLAKE} \\
\cmidrule(lr){3-4} \cmidrule(lr){5-5}
& & \textbf{real} & \textbf{synthetic} & \textbf{score} \\
\midrule
\multirow{2}{*}{Idefics2}
& LoRA & 0.8525 & 0.8489 & 0.7509 \\
& SWAG & 0.8595 & 0.8621 & 0.7488\\
\midrule
\multirow{2}{*}{Qwen2-VL}
& LoRA & 0.9012 & 0.9154 & 0.7502 \\
& SWAG & 0.9107 & 0.9286 & 0.7320 \\
\midrule
\multirow{2}{*}{DeepSeek-VL2}
& LoRA & 0.8984 & 0.8556 & 0.6619 \\
& SWAG & 0.8701 & 0.8329 & -- \\
\bottomrule
\end{tabular}}
\captionof{table}{\textbf{Target-model VQA performance} on the real and synthetic subsets of PFL-DocVQA and on SLAKE. We report normalized Levenshtein similarity~\cite{tito2024privacy} as the evaluation metric.
% We do not include DeepSeek-VL2 in our MedVQA experiments due to its low performance.
}
\label{tab:target_model_vqa}
\end{minipage}
\hfill
\begin{minipage}[t]{0.51\textwidth}
\centering
\small
\resizebox{\linewidth}{!}{%
\begin{tabular}{lcccccc}
\toprule
\multirow{3}{*}{\textbf{Model}} 
& \multicolumn{4}{c}{\textbf{PFL-DocVQA}} 
& \multicolumn{2}{c}{\textbf{SLAKE}} \\
\cmidrule(lr){2-5} \cmidrule(lr){6-7}
& \multicolumn{2}{c}{\textbf{real}} 
& \multicolumn{2}{c}{\textbf{synthetic}} 
& \multicolumn{2}{c}{\textbf{}} \\
\cmidrule(lr){2-3} \cmidrule(lr){4-5} \cmidrule(lr){6-7}
& \textbf{IN} & \textbf{OUT} 
& \textbf{IN} & \textbf{OUT} 
& \textbf{IN} & \textbf{OUT} \\
\midrule
Idefics2     & 0.8760 & 0.8523 & 0.9020 & 0.8489 & 0.7616 & 0.7509 \\
Qwen2-VL     & 0.9237 & 0.9012 & 0.9377 & 0.9154 & 0.7522 & 0.7502 \\
DeepSeek-VL2 & 0.8980 & 0.8984 & 0.9088 & 0.8556 & \multicolumn{2}{c}{--} \\
\bottomrule
\end{tabular}%
}
\captionof{table}{\textbf{Target-model performance on PFL-DocVQA and SLAKE.} We report results for the IN and OUT splits. The similar performance across the two splits indicates no clear sign of overfitting.}
\label{tab:target_model_in_out}
\end{minipage}
\end{table*}

Table~\ref{tab:target_model_vqa} reports the downstream VQA performance of the fine-tuned target models on PFL-DocVQA (real and synthetic subsets) and SLAKE, evaluated using normalized Levenshtein similarity~\citep{tito2024privacy}. All three VLMs achieve strong task performance,confirming that the resulting checkpoints are realistic target models for our MIA evaluation. To further verify that the models are not in a degenerate overfitting regime (which would render membership inference trivially easy and unrepresentative of practical privacy risk), Table~\ref{tab:target_model_in_out} compares performance on the IN (training member) and OUT (held-out non-member) splits. The gap between the two splits is small across all models and datasets, indicating that the models generalize well and that any membership signal exploited by the MIA arises from subtle distributional cues rather than gross memorization.

\begin{table}[h]
\begin{center}
\begin{small}
\small
\resizebox{\linewidth}{!}{%
\begin{tabular}{cccc}
\toprule
 & \textbf{Idefics2} & \textbf{Qwen2-VL} & \textbf{DeepSeek-VL2} \\
\midrule 
Checkpoint & \text{Qwen/Qwen2-VL-7B-Instruct} & \text{HuggingFaceM4/idefics2-8b} & \text{deepseek-ai/deepseek-vl2-tiny}\\
Optimizer & AdamW & AdamW & AdamW \\
Learning rate & 1e-5 & 1e-4 & 1e-4 \\
Batch size & 2 & 1 & 1 \\
Accum. steps & 16 & 16 &  16\\
Epoch & 3 & 3 & 3 \\
Warmup & 0.03 & 0.03 & 0.03 \\
Precision & fp16 & bf16 & bf16 \\
\midrule
LoRA $r$ & 4 & 4 & 4 \\
LoRA $\alpha$ & 8 & 8 & 8 \\
LoRA Dropout & 0.2 & 0.2 & 0.1\\
LoRA num. layers & 14 &  14 &  14\\
\midrule
SWAG learning rate & 5e-6 & 3e-5 & 3e-5 \\
SWAG start & 0.8 & 0.6 & 0.6 \\
SWAG frequency & 0.05 & 0.05 & 0.05 \\
SWAG max snapshots $M$ & 20 & 20 & {20} \\
SWAG anneal strategy & linear & linear & {linear} \\
SWAG schedule & constant & constant & constant \\
SWAG variance scale $s$ & 2.0 & 2.0 & {2.0} \\
SWAG samples $K$ & 15 & 15 & {15} \\
\bottomrule
\end{tabular}
}
\end{small}
\end{center}
\caption{\textbf{Training hyperparameters} used for each model throughout the experiments.}
\label{tab:training_hps}
\end{table}

\subsection{SWAG Training}
We adapt the official SWAG-LoRA implementation~\citep{onal2024gaussian}. Because we operate in the low-data fine-tuning regime, we modify the standard SWAG training~\cite{maddox2019simple} recipe in two ways. First, we begin SWAG collection shortly after the validation loss plateaus rather than waiting for full convergence. Second, we increase the snapshot collection frequency to obtain a more diverse posterior approximation within the limited training budget. The learning rate is annealed linearly from the value at the SWAG starting point to a target rate approximately $10\times$ smaller than the peak learning rate. 
Following~\citet{onal2024gaussian}, we retain $M=20$ snapshots and, at inference, draw $K=15$ samples from the approximate posterior with a scale factor of $\kappa=2.0$ throughout all main experiments. As we show in Appendix~\ref{app:persample_threshold}, tuning $\kappa$ can further improve both calibration and attack performance. Therefore, the results reported in Section~\ref{sec:results} conservative with respect to this hyperparameter. The same $K=15$ is used for MC Dropout to ensure a fair comparison.

\subsection{Baseline Implementations}
\label{app:attack_implementations}
\noindent\textbf{Reference-model attacks.} Following~\cite{zhang-etal-2024-order}, LiRA~\citep{carlini2022membership} and RMIA~\citep{zarifzadeh2023low} are trained on randomly sampled half-subsets of the shadow dataset, using the same fine-tuning recipe (LoRA configuration, optimizer, schedule, and number of epochs) as the target model. This ensures that the shadow models faithfully approximate the target's training distribution, which is the central assumption underlying both attacks.

\noindent\textbf{QMIA~\cite{bertran2023scalable}}. For the DocVQA experiments, the quantile regressor used by QMIA is built on a DiT~\citep{li2022dit} backbone and trained on the full shadow split for $30$ epochs with document-specific augmentations. We checkpoint after each epoch and retain the snapshot with the lowest validation loss on a held-out subset. The regressor predicts $N_q$ quantiles placed on a log-spaced grid concentrated in the upper tail, where MIA signal is strongest. Rather than assuming a Gaussian conditional, the regressor is trained with the pinball loss to directly match many quantile levels, yielding a flexible non-parametric model of the score distribution given the image.

\noindent\textbf{LLM-based attacks.} For Min-K\% Prob~\citep{shi2024detecting} and Max-Renyi-K\% Prob~\citep{li2024membership}, we use the authors' official implementations.\footnote{Min-K\% Prob: \url{https://github.com/swj0419/detect-pretrain-code}; Max-Renyi-K\% Prob: \url{https://github.com/LIONS-EPFL/VL-MIA}.} We report the best result obtained by sweeping over the recommended value range of $K$.

\section{Additional Results}
\label{app:results}

\subsection{Global Threshold}

We report results across three MIA settings: \emph{sample-level} on MedVQA (Table~\ref{tab:results_sample_level_medvqa}), and \emph{multi-sample} (Table~\ref{tab:result_multi_sample_docvqa}) and \emph{image-only} (Table~\ref{tab:result_image_only_docvqa}) on DocVQA.

As shown in Table~\ref{tab:results_sample_level_medvqa}, all MIAs achieve only limited performance on MedVQA, with AUC values close to random (50\%) and low TPR at small FPR thresholds. We attribute this to the relatively constrained nature of the task: MedVQA questions exhibit limited linguistic diversity and target a domain that modern VLMs handle competently, leaving little signal in the loss distribution to distinguish members from non-members. Despite this difficulty, \attackName{} consistently outperforms prior baselines on Qwen2-VL across all metrics, improving AUC by 1.3 points over the strongest competitor (RMIA) and nearly doubling TPR@3\%FPR (5.46 vs.\ 3.55).

\begin{table*}[h]
\centering

\begin{minipage}[t]{0.58\textwidth}
\vspace{0pt}
\centering
\small
\resizebox{\linewidth}{!}{%
\begin{tabular}{lcccccc}
\toprule
\multirow{3}{*}{\textbf{Method}} 
& \multicolumn{3}{c}{\textbf{Idefics2}} 
& \multicolumn{3}{c}{\textbf{Qwen2-VL}} \\
\cmidrule(lr){2-4} \cmidrule(lr){5-7}
& \multicolumn{2}{c}{\textbf{TPR@FPR}} 
& \multirow{2}{*}{\textbf{AUC}}
& \multicolumn{2}{c}{\textbf{TPR@FPR}} 
& \multirow{2}{*}{\textbf{AUC}} \\
\cmidrule(lr){2-3} \cmidrule(lr){5-6}
& \textbf{3\%} & \textbf{5\%} & 
& \textbf{3\%} & \textbf{5\%} & \\
\midrule
Attack-P & 2.99 & 5.00 & 53.27 & 1.92 & 4.03 & 52.10 \\
QMIA ($N=1$) & 2.75 & 5.00 & 53.27 & 2.75 & 5.00 & 53.27 \\
LiRA ($N=4$)$^\dagger$ &  4.15 & 6.06 & 56.88 & 2.14 & 4.06 & 55.35 \\
RMIA ($N=4$)$^\dagger$ & \multicolumn{3}{c}{--} & 3.55 & 4.71 & 56.35 \\
\attackName{} ($N=1$)$^\dagger$ & \textbf{6.22} & \textbf{7.87} & \textbf{58.29} & \textbf{5.46} & \textbf{8.42} & \textbf{57.66} \\
\bottomrule
\end{tabular}%
}
\captionof{table}{\textbf{Sample-level MIA results on Medical VQA for Idefics2 and Qwen2-VL.} We report TPR at 3\% and 5\% FPR, together with AUC. Best values are shown in \textbf{bold}.}
\label{tab:results_sample_level_medvqa}
\end{minipage}
\hfill
\begin{minipage}[t]{0.41\textwidth}
\vspace{0pt}
\centering
\small
\resizebox{\linewidth}{!}{%
\begin{tabular}{lcccc}
\toprule
\multirow{2}{*}{\textbf{Method}}
& \multicolumn{2}{c}{\boldmath$\alpha=0.03$}
& \multicolumn{2}{c}{\boldmath$\alpha=0.05$} \\
\cmidrule(lr){2-3} \cmidrule(lr){4-5}
& \textbf{FPR} & \textbf{TPR}
& \textbf{FPR} & \textbf{TPR} \\
\midrule
Attack-P                & 0.00  & 0.00  & 0.00  & 0.67  \\
QMIA ($N{=}1$)          & 0.33  & 7.67  & 1.33  & 10.33 \\
\attackName{} ($N{=}1$) & 12.67 & 61.67 & 17.33 & 66.00 \\
\bottomrule
\end{tabular}%
}
\captionof{table}{\textbf{Per-sample calibrated MIA at $\alpha\in\{0.03,0.05\}$ on Qwen2-VL/DocVQA.} FPR measures calibration; TPR measures power. \attackName{} is the only method producing a meaningful test under this protocol.}
\label{tab:mia_alpha_results}
\end{minipage}

\end{table*}

\subsection{Per-sample Calibrated Threshold}
\label{app:persample_threshold}
For methods admitting cheap null estimation, we additionally report results under the per-sample calibrated protocol: for each target $\mathbf{x}$, we reject at significance level $\alpha$ using a threshold $\tau(\mathbf{x})$ fit to a held-out non-member population, controlling the per-sample FPR at $\alpha$. This applies to \attackName{} and to the population-based baselines (Attack-P, QMIA); reference-based baselines (LiRA, Attack-R, RMIA) are omitted for the reasons given in Section~\ref{sec:evaluation_protocol}. We report the realized FPR against the nominal $\alpha$ (calibration) and the realized TPR (power) in Table~\ref{tab:mia_alpha_results}.
Population-based baselines calibrate near $\alpha$ but achieve near-zero power. \attackName{} trades modest over-rejection for substantially higher power, and is the only method producing a meaningful test under this protocol.

\subsection{Computational Cost}
Table~\ref{tab:wallclock_time} reports a wall-clock time breakdown on Qwen2-VL. All procedures are executed sequentially on a single NVIDIA A40 GPU (45\,GB) without parallelization, with a per-sample batch size of one during inference.
LiRA's cost comprises training and scoring $N{=}4$ independent reference models; RMIA additionally trains a validation model and scores a population set. \attackName{} requires training a single Bayesian reference model (including the SWAG collection phase) and scoring $K{=}15$ posterior samples at inference.
% \begin{table}[h]
% \centering
% \caption{\textbf{Wall-clock time (GPU-hours)} for reference-based attacks on Qwen2-VL. All methods are trained for one epoch on a single GPU using the hyperparameters in Table~\ref{tab:training_hps}. \attackName{} draws $K{=}15$ posterior samples at inference.}

% \label{tab:wallclock_time}
% \begin{tabular}{lccc}
% \toprule
%  & \attackName $(N=1)$ & LiRA $(N=4)$ & RMIA $(N=4)$ \\
% \midrule
% Training  & 1.30 & 5.66 & 7.11 \\
% Inference & 7.16 & 3.36 & 5.01 \\
% \midrule
% Total     & 8.46 & 9.02 & 12.12 \\
% Training speedup & --- & $4.4\times$ & $5.5\times$ \\
% \bottomrule
% \end{tabular}
% \end{table}

\begin{table}[h]
\centering
\begin{tabular}{lcccc}
\toprule
 & \multicolumn{2}{c}{\attackName $(N=1)$} & \multirow{2}{*}{\makecell[c]{LiRA \\ $(N=K=4)$}{}} & \multirow{2}{*}{\makecell[cc]{RMIA \\ $(N=K=4)$}{}} \\
\cmidrule(lr){2-3}
 & $K=15$ & $K=4$ &  &  \\
\midrule
Training        & 1.30 & 1.30 & 5.66 & 7.11 \\
Inference       & 7.16 & 3.03 & 3.36 & 5.01 \\
\midrule
Total           & 8.46 & 4.33 & 9.02 & 12.12 \\
Training speedup & \multicolumn{2}{c}{---} & $4.4\times$ & $5.5\times$ \\
\bottomrule
\end{tabular}
\vspace{5pt}
\caption{\textbf{Wall-clock time (GPU-hours)} for reference-based attacks on Qwen2-VL. All methods are trained for one epoch on a single GPU using the hyperparameters in Table~\ref{tab:training_hps}. By default, \attackName{} draws $K{=}15$ posterior samples at inference; we also report the matched-ensemble setting $K{=}4$.}
\label{tab:wallclock_time}
\end{table}

The training phase dominates the savings: \attackName{} trains a single fine-tuning run in place of multiple independent shadow models, reducing training time by $4.4\times$ relative to LiRA and $5.5\times$ relative to RMIA. This advantage is independent of $K$, since posterior sampling adds no training cost.

At the default $K{=}15$, inference is higher for \attackName{} because each target is scored under all $K$ posterior samples, whereas LiRA and RMIA query only $N{=}4$ pre-trained models. Even so, the total wall-clock time remains competitive with LiRA and $1.4\times$ lower than RMIA, confirming that the Bayesian posterior amortizes the reference ensemble without introducing an overall bottleneck.

At $K{=}4$, \attackName{} becomes cheaper on \emph{both} axes: inference drops to 3.03 GPU-h (below LiRA’s 3.36), and the total falls to 4.33 GPU-h, a $2.1\times$ reduction over LiRA and $2.8\times$ over RMIA. This is the matched-ensemble setting of Table~\ref{tab:k_matched_ablation}, where \attackName{} already matches LiRA$(N{=}4)$ in attack performance. \attackName{} therefore attains the same accuracy as a four-model shadow ensemble at roughly half the total cost, and recovers the full low-FPR gains in the main tables by scaling $K$ at no additional training cost.

\begin{table*}[t]
\centering
\small
\resizebox{\textwidth}{!}{
\begin{tabular}{lcccccccccccccc}
\toprule
\multirow{3}{*}{\textbf{Method}} 
& \multirow{3}{*}{\textbf{$K$}} 
& \multicolumn{4}{c}{\textbf{Idefics2}} 
& \multicolumn{4}{c}{\textbf{Qwen2VL}} 
& \multicolumn{4}{c}{\textbf{DeepSeek-VL2}} \\
\cmidrule(lr){3-6} \cmidrule(lr){7-10} \cmidrule(lr){11-14}
& 
& \multicolumn{2}{c}{$\mathrm{D}_{\mathrm{real}}$}
& \multicolumn{2}{c}{$\mathrm{D}_{\mathrm{\textcolor{red}{synth}}}$}
& \multicolumn{2}{c}{$\mathrm{D}_{\mathrm{real}}$}
& \multicolumn{2}{c}{$\mathrm{D}_{\mathrm{\textcolor{red}{synth}}}$}
& \multicolumn{2}{c}{$\mathrm{D}_{\mathrm{real}}$}
& \multicolumn{2}{c}{$\mathrm{D}_{\mathrm{\textcolor{red}{synth}}}$} \\
\cmidrule(lr){3-4} \cmidrule(lr){5-6}
\cmidrule(lr){7-8} \cmidrule(lr){9-10}
\cmidrule(lr){11-12} \cmidrule(lr){13-14}
& 
& \textbf{3\%} & \textbf{5\%}
& \textbf{3\%} & \textbf{5\%}
& \textbf{3\%} & \textbf{5\%}
& \textbf{3\%} & \textbf{5\%}
& \textbf{3\%} & \textbf{5\%}
& \textbf{3\%} & \textbf{5\%} \\
\midrule
\multicolumn{11}{l}{\textbf{Population-data}}\\

Min-K\%Prob~\cite{shi2024detecting} & -- & 5.67 & 11.33 & 10.33 & 16.00 & 15.00 & 22.00 & 11.33 & 14.33 & 2.67 & 4.00 & 14.00 & 20.00\\
MaxRenyi-K\%~\cite{li2024membership} & -- &  11.33 & 13.67 & 9.67 & 13.00 & 14.67 & 17.67 & 9.67 & 14.67 & \secondbest{13.00} & \secondbest{14.33} & 7.00 & 10.67\\
Attack-P~\cite{ye2022enhanced} & 1200 & 4.00 & 11.33 & 8.67 & 15.33 & 13.67 & 18.33 & 9.00 & 11.33 & 0.67 & 4.33 & 13.00 & 20.67 \\
Q-MIA~\cite{bertran2023scalable}($N=1$) & 1200 & 5.67 & 12.67 & 10.67 & 15.00 & 16.00 & 21.33 & 10.00 & 14.33 & 2.67 & 4.33 & 14.33 & 19.67 \\
\midrule
\multicolumn{11}{l}{\textbf{Reference-model}}\\
% Attack-R~\cite{ye2022enhanced}($N=8$) & 2 & & & & & & & & & & & & \\
LiRA~\cite{carlini2022membership}($N=2$) & 2 & 13.00 & 18.67 & 5.33 & 10.33 & 10.00 & 14.67 & 3.33 & 6.67 & 3.00 & 3.67 & 7.67 & 11.00 \\
LiRA~\cite{carlini2022membership}($N=4$) & 4 & 28.67 & 38.33 & 22.67 & 28.33 & \best{40.67} & \best{47.33} & 8.00 & 8.33 & 4.00 & 6.00 & 6.67 & 12.00 \\
LiRA$\mathbf{\ast}$~\cite{carlini2022membership}($N=4$) & 4 & 24.33 & 30.00 & \secondbest{25.67} & \secondbest{30.33} & 30.67 & 36.00 & \secondbest{12.33} & \best{28.00} & 0.67 & 2.33 & 3.33 & 8.67 \\
% LiRA~\cite{carlini2022membership}($N=8$) & 8 & {39.67} & {49.67} & {36.33} & {41.00} & {52.67} & {58.67} & 9.33 & 19.67 & 7.33 & 10.00 & 7.67 & 9.00 \\
RMIA~\cite{zarifzadeh2023low}($N=2$) & 2 & 31.67 & 38.33 & 18.33 & 26.33 & 36.67 & 42.00 & 7.33 & 10.33 & 1.67 & 8.67 & \secondbest{19.67} & \secondbest{25.33} \\
RMIA~\cite{zarifzadeh2023low}($N=4$) & 4 & \secondbest{37.33} & \secondbest{42.00} & 15.67 & 28.33 & 36.00 & 36.00 & 9.33 & 13.67 & 5.33 & 8.67 & 17.33 & 25.00 \\
% RMIA~\cite{zarifzadeh2023low}($N=8$) & 8 & 28.00 & 39.33 & 20.67 & 28.67 & 40.00 & 47.00 & 9.00 & 18.67 & 1.67 & {11.00} & 19.00 & 27.00 \\
\midrule
\attackName{} ($N=1$) & 15 & \best{38.00} & \best{44.67} & \best{31.67} & \best{38.00} & \secondbest{38.33} & \secondbest{45.67} & \best{16.00} & \secondbest{25.00} & \best{14.00} & \best{17.33} & \best{24.33} & \best{34.67}\\
\bottomrule
\end{tabular}
}
\caption{\textbf{Multi-sample MIA on DocVQA.} TPR at 3\% and 5\% FPR on the real ($\mathrm{D}_{\mathrm{real}}$) and synthetic ($\mathrm{D}_{\textcolor{red}{\mathrm{synth}}}$) subsets. LiRA$^{\ast}$ indicates its variant using a global variance. Min-K\%Prob and MaxRenyi-K\% results are reported with best hyperparameters. Best values are shown in \best{red}; second-best values are shown in \secondbest{blue}.}
\label{tab:result_multi_sample_docvqa}
\end{table*}

\begin{table*}[t]
\centering
\small
\resizebox{\textwidth}{!}{
\begin{tabular}{lcccccccccccccc}
\toprule
\multirow{3}{*}{\textbf{Method}} 
& \multirow{3}{*}{\textbf{$K$}} 
& \multicolumn{4}{c}{\textbf{Idefics2}} 
& \multicolumn{4}{c}{\textbf{Qwen2VL}} 
& \multicolumn{4}{c}{\textbf{DeepSeek-VL2}} \\
\cmidrule(lr){3-6} \cmidrule(lr){7-10} \cmidrule(lr){11-14}
& 
& \multicolumn{2}{c}{$\mathrm{D}_{\mathrm{real}}$}
& \multicolumn{2}{c}{$\mathrm{D}_{\mathrm{\textcolor{red}{synth}}}$}
& \multicolumn{2}{c}{$\mathrm{D}_{\mathrm{real}}$}
& \multicolumn{2}{c}{$\mathrm{D}_{\mathrm{\textcolor{red}{synth}}}$}
& \multicolumn{2}{c}{$\mathrm{D}_{\mathrm{real}}$}
& \multicolumn{2}{c}{$\mathrm{D}_{\mathrm{\textcolor{red}{synth}}}$} \\
\cmidrule(lr){3-4} \cmidrule(lr){5-6}
\cmidrule(lr){7-8} \cmidrule(lr){9-10}
\cmidrule(lr){11-12} \cmidrule(lr){13-14}
& 
& \textbf{3\%} & \textbf{5\%}
& \textbf{3\%} & \textbf{5\%}
& \textbf{3\%} & \textbf{5\%}
& \textbf{3\%} & \textbf{5\%}
& \textbf{3\%} & \textbf{5\%}
& \textbf{3\%} & \textbf{5\%} \\
\midrule
\multicolumn{11}{l}{\textbf{Population-data}}\\
Min-K\%Prob~\cite{shi2024detecting} & -- & 6.00 & 8.00 & 4.00 & 7.33 & 7.67 & 12.00 & 3.00 & 5.33 & 4.00 & 7.67 & 7.00 & 10.00\\
MaxRenyi-K\%~\cite{li2024membership} & -- & 10.00 & 14.00 & 5.33 & 5.33 & 9.00 & 9.67 & 3.33 & 3.67 & \best{13.67} & \best{17.67} & 10.33 & 11.33\\
Attack-P~\cite{ye2022enhanced} & 1200 & 2.33 & 5.33 & 2.33 & 2.33 & 6.33 & 12.67 & 0.00 & 0.00 & 0.00 & 6.00 & 7.00 & 7.00 \\
Q-MIA~\cite{bertran2023scalable}($N=1$) & 1200 & 4.67 & 11.67 & 4.33 & 8.33 & 3.67 & 6.00 & \best{10.00} & \best{14.33} & 7.67 & 11.33 & 10.00 & \secondbest{16.67} \\
\midrule
\multicolumn{11}{l}{\textbf{Reference-model}}\\
LiRA~\cite{carlini2022membership}($N=2$) & 2 & 4.33 & 10.67 & 4.00 & 7.67 & 6.00 & 12.33 & 2.67 & 3.00 & 1.33 & 3.67 & 4.00 & 4.67 \\
LiRA~\cite{carlini2022membership}($N=4$) & 4 & \secondbest{11.33} & 15.67 & 5.67 & 9.00 & 6.33 & 18.00 & 2.00 & 4.33 & 4.00 & 6.00 & 5.67 & 7.00 \\
LiRA$\mathbf{\ast}$~\cite{carlini2022membership}($N=4$) & 4 & 10.67 & 12.00 & 4.33 & 8.33 & 6.00 & 9.67 & 3.00 & 7.67 &  0.67 & 4.33 & 3.00 & 7.67 \\
RMIA~\cite{zarifzadeh2023low}($N=2$) & 2 & 7.00 & \secondbest{33.00} & 5.67 & 16.00 & 23.67 & 27.33 & 4.67 & 9.33 & 2.00 & 2.00 & 9.00 & 14.33\\
RMIA~\cite{zarifzadeh2023low}($N=4$) & 4 & 7.33 & 30.00 & \secondbest{13.33} & \secondbest{18.33} & \best{25.00} & \best{32.33} & 0.00 & 8.33 & 2.00 & 8.67 & \secondbest{10.33} & 16.00\\
\midrule
\attackName{} ($N=1$) & 15 & \best{33.00} & \best{36.00} & \best{21.00} & \best{24.67} & \secondbest{24.67} & \secondbest{30.33} & \secondbest{7.67} & \secondbest{10.67} & \secondbest{11.00} & \secondbest{13.33} & \best{21.33} & \best{29.00} \\
\bottomrule
\end{tabular}
}
\caption{\textbf{Image-only MIA on DocVQA.} TPR at 3\% and 5\% FPR on the real ($\mathrm{D}_{\mathrm{real}}$) and synthetic ($\mathrm{D}_{\textcolor{red}{synth}}$) subsets. Best values are shown in \best{red}; second-best values are shown in \secondbest{blue}.}
\label{tab:result_image_only_docvqa}
\end{table*}

\section{Attack Baselines}
\label{app:baselines}

In this section, we provide additional details on the baseline attacks compared in our evaluation. We group them by the type of side information used to estimate the null distribution $p(S \mid \mathcal{H}_0)$: \emph{Population-Data attacks}, which rely solely on auxiliary samples from $\mathcal{D}$, and \emph{Reference-Model attacks}, which additionally train shadow models. Table~\ref{tab:mia_setting_summary} summarizes their key properties.

\subsection{Population-Data Attacks}

\noindent\textbf{Attack P~\cite{ye2022enhanced}.}
Attack P estimates the null distribution \emph{marginally} from the target model's behavior on population data, without conditioning on the covariates of the target sample $\mathbf{x}$. Concretely, the adversary samples $\{\mathbf{x}_i\}_{i=1}^m \sim \mathcal{D}$ independently from $\mathbf{x}$, computes the scores $\{S(\mathbf{x}_i, \mathcal{F})\}_{i=1}^m$ on the target model, and uses them to approximate $p(S \mid \mathcal{H}_0)$. The decision rule thresholds $S(\mathbf{x}, \mathcal{F})$ at the $\alpha$-percentile $c_\alpha(\mathcal{F})$ of this distribution. Because a single threshold is shared across all targets, the FPR guarantee $\alpha$ holds only at the population level, and the resulting TPR@FPR reflects average-case leakage over $\mathbf{x} \sim \mathrm{D}$ (members) or $\mathbf{x} \sim D \setminus \mathrm{D}$ (non-members). Attack P avoids training shadow models, but accurate calibration requires sufficient population data to estimate the $\alpha$-quantile tightly and minimize the gap between the nominal and empirical FPR.

\noindent\textbf{QMIA~\cite{bertran2023scalable}.}
Quantile regression MIA replaces the marginal null estimate of Attack P with a \emph{conditional} one: rather than using a single threshold for all targets, QMIA trains a regressor that predicts the $\alpha$-quantile of $p(S \mid \mathcal{H}_0(\mathbf{x}))$ directly from the covariates of $\mathbf{x}$. The decision rule applies a per-target threshold $\tau(\mathbf{x})$ obtained from this regressor. Two practical advantages distinguish QMIA from shadow-model attacks: it requires training only a single auxiliary model (the quantile regressor), and the regressor's architecture need not match that of the target model, so no knowledge of the target's training algorithm is required. The underlying assumption, shared with most score-based MIAs, is that $\mathcal{F}$ systematically assigns higher confidence to training points than to fresh samples from the same distribution.
\begin{table}[t]
\begin{center}
\begin{small}
\resizebox{\textwidth}{!}{
\begin{tabular}{lcccccc}
    \toprule
     \textbf{Method} & \makecell{\textbf{Conditioned}\\\textbf{on $\mathrm{x}$}} & \makecell{\textbf{Decision}\\\textbf{Threshold}} & \makecell{\textbf{FPR}\\\textbf{Guarantee}} & \makecell{$p(S \mid \mathcal{H}_0)$\\\textbf{Estimate}} & \makecell{\textbf{Estimation}\\\textbf{Basis}} & \textbf{Metric} \\
    \midrule
    Attack-R~\cite{ye2022enhanced}     & Yes & $c_\alpha(\mathrm{x})$ & $\alpha(\mathrm{x})$  & Empirical              & Reference models & \multirow{3}{*}{TPR@lowFPR($\mathrm{x}$)} \\
    QMIA~\cite{bertran2023scalable}    & Yes & $c_\alpha(\mathrm{x})$ & $\alpha(\mathrm{x})$  & Quantile regression    & Population data  & \\
    BMIA~\cite{zhang2024membership}    & Yes & $c_\alpha(\mathrm{x})$ & $\alpha(\mathrm{x})$  & Laplace approximation  & Reference models & \\
    \midrule
    Attack-P~\cite{ye2022enhanced}     & No  & $c_{\alpha}$           & $\alpha$              & Empirical              & Population data  & \multirow{4}{*}{TPR@lowFPR} \\
    LiRA~\cite{carlini2022membership}  & Yes & $c$                    & No$^\ast$             & Gaussian approximation & Reference models & \\
    RMIA~\cite{zarifzadeh2023low}      & Yes & $c$                    & No$^\ast$             & Empirical              & Reference models & \\
    \attackName{}                      & Yes & $c$                    & No$^\ast$             & SWAG~\cite{maddox2019simple} & Reference models &  \\
    \bottomrule
\end{tabular}
}
\end{small}
\end{center}
\caption{\textbf{Comparison of existing MIAs by privacy-leakage regime, threshold construction, FPR control, null-distribution estimator, and evaluation metric.} \emph{Conditioned on $\mathrm{x}$} indicates whether the null distribution is estimated per target sample or marginally across the population. The top block reports TPR at a global FPR averaged across targets, while the middle block reports TPR at a per-target FPR $\alpha(\mathrm{x})$ aggregated across the test population. $^\ast$LiRA, RMIA, and \attackName{} fit per-target null distributions and could in principle yield a threshold $c_\alpha(\mathrm{x})$ with an $\alpha(\mathrm{x})$-FPR guarantee, but their original protocols sweep a global threshold $c$ and do not provide per-sample FPR control.}
\label{tab:mia_setting_summary}
\end{table}

\subsection{Reference-Model Attacks}

\noindent\textbf{LiRA~\cite{carlini2022membership}.}
The Likelihood Ratio Attack scores membership by comparing the target model's score on $\mathbf{x}$ against the distribution of scores obtained from reference models that did \emph{not} see $\mathrm{x}$. In the offline variant we use, the adversary trains $N$ reference models $\{\mathcal{F}^i\}_{i=1}^N$ on datasets $\mathrm{D}^i \sim \mathcal{D}^n$ with $\mathbf{x} \notin \mathrm{D}^i$, fits a Gaussian to the reference scores $\{S(\mathbf{x}, \mathcal{F}^i)\}_{i=1}^N$ to approximate $p(S \mid \mathcal{H}_0(\mathbf{x}))$, and computes the attack score $\Lambda(\mathbf{x}, \mathcal{F}) = 1 - \Pr[S(\mathbf{x}, \mathcal{F}) \mid \mathcal{H}_0(\mathbf{x})]$. By normalizing for record-specific score variability, LiRA achieves stronger separation than fixed-threshold attacks, particularly in the low-FPR regime emphasized by~\cite{carlini2022membership}. In the standard evaluation protocol, however, a \emph{single} threshold $c$ is swept across all targets, so the reported TPR@low-FPR controls FPR only at the population level. While the parametric per-target null in principle admits a per-target threshold $c_\alpha(\mathbf{x})$, accurately estimating tail quantiles of the OUT distribution per record is statistically noisy with limited reference models, which is why the original work reports global-threshold ROC.

\noindent\textbf{RMIA~\cite{zarifzadeh2023low}.}
RMIA strengthens reference-model attacks via a \emph{relativity} principle: rather than testing whether $\mathbf{x}$'s score is extreme under its own null world in isolation, it asks whether $\mathbf{x}$ is more extreme under $\mathcal{H}_0(\mathbf{x})$ than randomly drawn population points $\mathbf{x}'$ are under their own null worlds $\mathcal{H}_0(\mathbf{x}')$. Given reference models $\{\mathcal{F}^i\}_{i=1}^N$ and population samples $\{\mathbf{x}_j'\}_{j=1}^k$, RMIA computes a pairwise likelihood ratio
\begin{equation}
\textsc{LR}_{\mathcal{F}}(\mathbf{x}, \mathbf{x}_j') = \left(\frac{S(\mathbf{x}, \mathcal{F})}{\tfrac{1}{N}\sum_{i} S(\mathbf{x}, \mathcal{F}^i)}\right) \Big/ \left(\frac{S(\mathbf{x}_j', \mathcal{F})}{\tfrac{1}{N}\sum_{i} S(\mathbf{x}_j', \mathcal{F}^i)}\right),
\end{equation}
and aggregates these into the membership score $\Lambda(\mathrm{x}, \mathcal{F}) = \Pr_{\mathbf{x}'}[\textsc{LR}_{\mathcal{F}}(\mathbf{x}, \mathbf{x}') \geq \gamma]$ for a fixed $\gamma$. Comparing $\mathbf{x}$ against many population points reduces the ambiguity caused by sample-specific difficulty, as some non-members are easy and some members are hard, and amplifies the test statistic even with few reference models. As with LiRA, the standard evaluation uses a single decision threshold across all targets, providing only population-level FPR control rather than per-target guarantees.

\noindent\textbf{Attack R~\cite{ye2022enhanced}.}
Attack R is the per-sample analogue of Attack P, structurally similar to offline LiRA but with a different decision rule and evaluation protocol. The adversary trains reference models $\{\mathcal{F}^i\}_{i=1}^N$ on datasets excluding $\mathbf{x}$, evaluates them on the target to obtain $\{S(\mathbf{x}, \mathcal{F}^i)\}_{i=1}^N$, and uses these scores to approximate $p(S \mid \mathcal{H}_0(\mathbf{x}))$. The decision threshold is set to the $\alpha$-percentile of this per-target null, yielding a personalized threshold $c_\alpha(\mathbf{x})$ for every $\mathbf{x}$. Unlike LiRA and RMIA, Attack R provides a per-target FPR guarantee $\alpha(\mathbf{x})$ by construction. As in LiRA and RMIA, reference datasets $\{\mathrm{D}^i\}_{i=1}^N$ can be sampled offline and reused across targets so long as they exclude the evaluated sample.

\noindent\textbf{BMIA~\cite{zhang2024membership}.}
The Bayesian MIA closest to our work amortizes the cost of training many reference models by replacing the empirical reference-score distribution with a \emph{post-hoc Laplace posterior} over the target model's parameters~\cite{liu2026how}. Concretely, BMIA fits a Gaussian approximation to the posterior $p(\theta \mid \mathrm{D})$ around the trained weights using a curvature estimate (typically the diagonal or Kronecker-factored Fisher), then propagates this parameter uncertainty to the score space to obtain a per-target null $p(S \mid \mathcal{H}_0(\mathrm{x}))$ without training additional reference models. Like Laplace-based Bayesian LoRA more broadly~\cite{yang2024bayesian}, however, the approach is restricted to classification tasks with a fixed output space, where the requisite curvature estimates remain tractable. We therefore do not compare to BMIA in our generative setting: the Laplace approximation is not directly applicable to open-ended generative models, whose output space is combinatorial and whose log-likelihood Hessian cannot be reduced to a fixed-size Fisher matrix in the same way.

\clearpage
% \newpage
% \input{checklist.tex}

\end{document}